\pdfoutput=1

\documentclass[11pt]{article}

\usepackage{emnlp2021}

\usepackage{times}
\usepackage{latexsym}

\usepackage[T1]{fontenc}

\usepackage[utf8]{inputenc}

\usepackage{microtype}

\usepackage{graphicx}
\usepackage{graphics}
\usepackage{hyperref}
\usepackage{url}
\usepackage{multirow}

\usepackage{colortbl}
\usepackage{makecell}

\usepackage{booktabs}
\usepackage{arydshln}
\usepackage{tabularx}
\newcolumntype{C}{>{\centering\arraybackslash}X}
\usepackage{footmisc}

\def\best{\bf\cellcolor[gray]{0.85}}
\def\secbest{\cellcolor[gray]{0.92}}
\definecolor{TableRed}{HTML}{800000}
\definecolor{mediumelectricblue}{rgb}{0.01, 0.31, 0.59}

\newcommand{\electricblue}[1]{\textcolor{mediumelectricblue}{#1}}
\usepackage{mdwlist}
\usepackage[list=true]{subcaption}
\usepackage{amsmath,amsfonts,amssymb}
\usepackage{pifont}
\newcommand{\cmark}{\textcolor{green}{\ding{51}}}%
\newcommand{\xmark}{\textcolor{red}{\ding{55}}}%

\newcommand{\method}{\textsc{DeepEx}}
\newcommand{\textspt}[1]{\texttt{#1}}
\newcommand{\comm}[1]{}

\newif\ifshowcomment
\showcommentfalse

\ifshowcomment
    \newcommand{\todo}[1]{\textcolor{yellow}{[TODO: #1]}}
    \newcommand{\dawn}[1]{\textcolor{purple}{[{Dawn: #1}]}}
    \newcommand{\jie}[1]{\textcolor{magenta}{[{Jie: #1}]}}
    \newcommand{\xiao}[1]{\textcolor{green}{[Xiao: #1]}}
    \newcommand{\zui}[1]{\textcolor{red}{[Zui: #1]}}
    \newcommand{\haoyun}[1]{\textcolor{blue}{[Haoyun: #1]}}
    \newcommand{\focus}[1]{\textcolor{orange}{#1}}
\else
    \newcommand{\todo}[1]{}
    \newcommand{\dawn}[1]{}
    \newcommand{\jie}[1]{}
    \newcommand{\xiao}[1]{}
    \newcommand{\zui}[1]{}
    \newcommand{\haoyun}[1]{}
    \newcommand{\focus}[1]{}
\fi

\title{Zero-Shot Information Extraction as a Unified Text-to-Triple Translation}

\author{Chenguang Wang$^*$, Xiao Liu$^\P$, Zui Chen$^\P$, Haoyun Hong$^\P$, Jie Tang$^\P$, Dawn Song$^*$ \\
$^*$UC Berkeley, $^\P$Tsinghua University \\
\texttt{\{chenguangwang,dawnsong\}@berkeley.edu}, \texttt{jietang@tsinghua.edu.cn} \\
\texttt{\{liuxiao21,chenzui19,honghy17\}@mails.tsinghua.edu.cn} \\
}

\begin{document}
\maketitle
\begin{abstract}
        We cast a suite of information extraction tasks into a text-to-triple translation framework. Instead of solving each task relying on task-specific datasets and models, we formalize the task as a translation between task-specific input text and output triples. By taking the task-specific input, we enable a task-agnostic translation by leveraging the latent knowledge that a pre-trained language model has about the task. We further demonstrate that a simple pre-training task of predicting which relational information corresponds to which input text is an effective way to produce task-specific outputs. This enables the zero-shot transfer of our framework to downstream tasks. We study the zero-shot performance of this framework on open information extraction (OIE2016, NYT, WEB, PENN), relation classification (FewRel and TACRED), and factual probe (Google-RE and T-REx). The model transfers non-trivially to most tasks and is often competitive with a fully supervised method without the need for any task-specific training. For instance, we significantly outperform the F1 score of the supervised open information extraction without needing to use its training set.\footnote{The code and datasets are available at \url{https://github.com/cgraywang/deepex}.}
\end{abstract}
\section{Introduction}
Information extraction refers to the task of automatically extracting structured information from unstructured resources, benefiting a wide range of applications such as information retrieval and knowledge base population. Information extraction covers a great variety of tasks in natural language processing (NLP), such as open information extraction and relation classification. For example, given a sentence ``Born in Glasgow, Fisher is a graduate of the London Opera Centre'', open information extraction seeks to extract (Fisher; Born in; Glasgow), and ``city\_of\_birth'' is predicted as the relation between a given pair of entities ``Fisher'' and ``Glasgow'' for relation classification. 

Most current approaches design task-specific pipelines for different information extraction tasks. Yet, this presents two limitations for information extraction. First, since most of the approaches employ a task-specific model, it is difficult to leverage a single pipeline to solve many tasks or adapt a model trained on one task to another without changing any task-specific modules. Second, those supervised state-of-the-arts are trained on task-specific corpora to predict from a fixed set of task-specific categories, which restricts their usability since additional labeled data is needed to specify any other classes. Such task-specific labeled data is scarce in information extraction. For example, the largest training set for open information extraction contains only 3,200 sentences~\cite{stanovsky2018supervised}. Motivated by this, we aim to solve information extraction tasks within the same framework in a task-agnostic setting.

\begin{figure*}[h]
    \centering
    \includegraphics[width=0.95\textwidth]{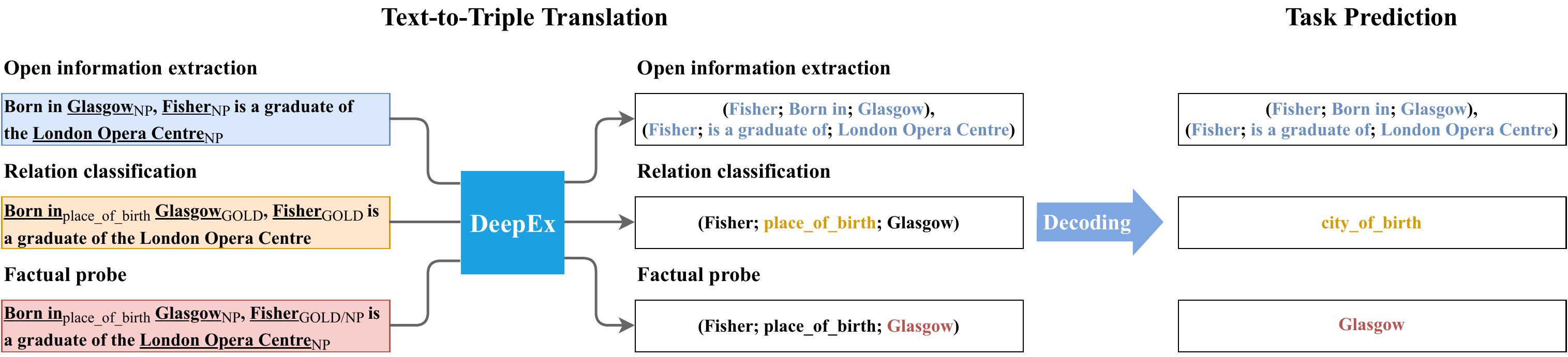}
    \caption{{\small Our \method\ translates between input text and output triples, and the output is then decoded into task predictions.}
      \label{fig:overview}}
\end{figure*}

In this paper, we propose a unified framework for information extraction. The basic idea is to treat every information extraction problem as a ``text-to-triple'' problem, i.e., translating input text to output triples. We successfully apply our framework to three information extraction tasks, greatly improving zero-shot performance on many datasets and sometimes even reaching competitiveness with the current state-of-the-art fully supervised approaches. Figure~\ref{fig:overview} shows how different information extraction tasks are handled within our framework. The framework encodes task priors in the input text and decodes the output triples to finally produce task predictions. We achieve this by leveraging the same translation process on all tasks, the only difference among tasks being the input encoding. This is in contrast with previous approaches using task-specific models and datasets. The design of the common translation module for all tasks is important: by leveraging the task priors encoded in the input text, we enable the zero-shot transfer of the general knowledge that a pre-trained LM has about the task. We demonstrate that a simple pre-training task of predicting which relational triple goes with which text on a task-agnostic corpus further enhances the zero-shot capabilities on all tasks. To the best of our knowledge, this is the first framework to handle a variety of information extraction tasks in a zero-shot setting.

Our contributions are summarized below.
\begin{enumerate}
    \item We introduce \method, a unified framework that solves information extraction tasks in a zero-shot setting. We cast information extractions as text-to-triple problems by incorporating the task priors in the input text and translating the input text to triples as output.
    \item We apply our framework to (\expandafter{\romannumeral1}) open information extraction, (\expandafter{\romannumeral2}) relation classification, and (\expandafter{\romannumeral3}) factual probe. In all tasks, we achieve competitive zero-shot performance to the current state-of-the-art including the fully supervised methods, and we achieve new state-of-the-art performance on open information extraction (OIE2016, WEB, NYT, and PENN) and factual probe (T-REx). For instance, our zero-shot approach significantly outperforms the supervised open information extraction by averaging 37.5 points in F1.
    \item We also show that our framework delivers more interpretable results while achieving comparable performance on all tasks, thanks to the transparency of the text-to-triple translation.
\end{enumerate}
\section{Method}
\label{sec:method}
We cast a suite of information extraction tasks into a text-to-triple translation framework. As shown in Figure~\ref{fig:overview}, input and output are designed in a format that is appropriate for a given task. The translation process takes the input text and produces triples as output. The decoding step generates task predictions from the output. In this section, we describe the input and output format, the translation, and the decoding process. We use open information extraction (OIE) as a running example in this section. For OIE, we are given a sentence and asked to extract triples. 

\begin{figure*}[h]
    \centering
    \includegraphics[width=1\textwidth]{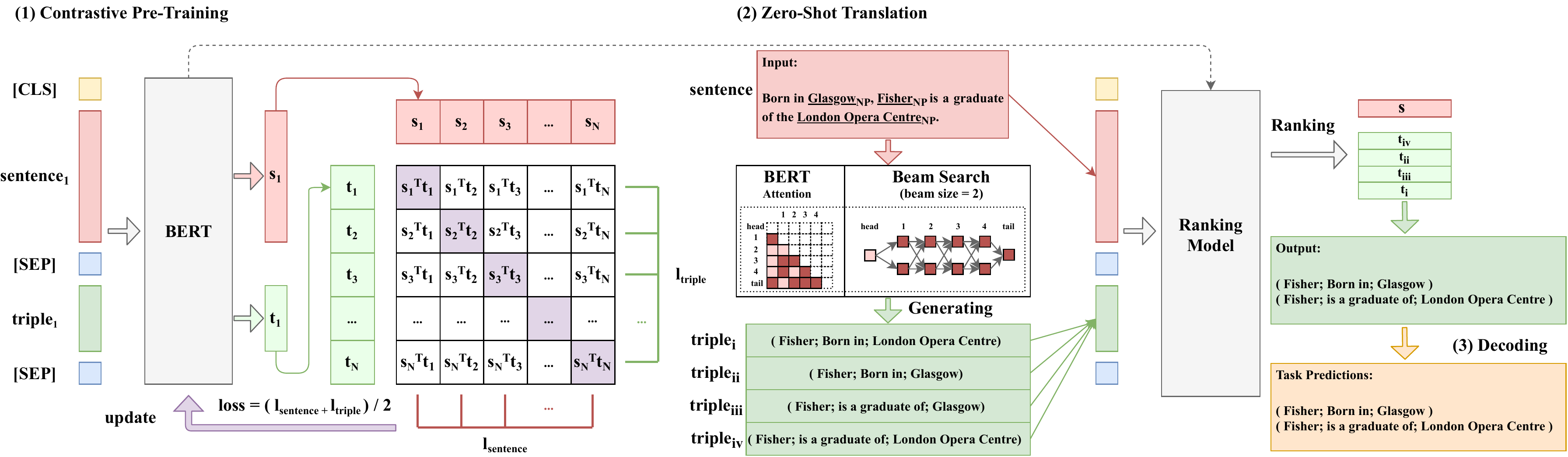}
    \caption{{\small Summary of our approach. The framework encodes task-relevant information in the input text and decodes the output triples to produce task predictions. The zero-shot translation first generates general information that a pre-trained language model has about the input, then ranks to find the output triples of interest to the task via a ranking model pre-trained on a task-agnostic relational corpus.}
      \label{fig:oie}}
\end{figure*}

\subsection{Input and Output Format}
\label{sec:inoutoie}
The input is a NP-chunked sentence, and the output is a set of triples. The NPs are encoded as task priors in the input. Below is an example.
\begin{enumerate*}
    \item[] {\bf Input} Born in \underline{Glasgow}$_{\rm NP}$, \underline{Fisher}$_{\rm NP}$ is a graduate of the \underline{London Opera Centre}$_{\rm NP}$.
    \item[] {\bf Output} (Fisher; Born in; Glasgow), (Fisher; is a graduate of; London Opera Centre).
\end{enumerate*}
${\rm NP}$ denotes the noun phrase. 

\subsection{Zero-Shot Translation}
\label{sec:gen}
We aim to translate the above input text to output triples. Information extraction tasks lack high-quality training data, therefore training an end-to-end supervised approach~\cite{paolini2021structured} is not feasible. Pre-trained language models (LM) (e.g., BERT~\cite{devlin2018bert} and GPT~\cite{brown2020language}) have demonstrated their zero-shot capabilities in a wide range of NLP tasks, thanks to the general information that they know about the tasks. 

We therefore propose a zero-shot translation process consisting of two steps: generating and ranking, as shown in Figure~\ref{fig:oie}. The generating stage produces general information about the task via pre-trained LMs, and the ranking stage looks for specific information about the task via a ranking model pre-trained on a task-agnostic corpus. 

\paragraph{Generating} The generating stage produces a set of candidate triples that contain general information about the task from pre-trained LMs. OIE is formulated as extracting a set of sequences in the input that are generally relevant to an argument pair (i.e., NP pair). We particularly use the attention scores in pre-trained LMs to measure the relevance between the sequence and the argument pair. 

We frame the process as a search problem. Given an argument pair (e.g., ``Fisher'' and ``London Opera Centre''), we aim to search for the sequences (e.g., ``is a graduate of'') with the largest attention scores connecting the pair. To compute a score for every possible sequence is computationally expensive, especially when the sequence length is large. Therefore the exhaustive search is intractable. We use beam search, which is an approximate strategy to explore the search space efficiently. Beam search maintains the $k$-best candidates. This means the time cost of beam search does not depend on the sequence length but the size of the beam and the average length of the candidates. The beam search starts with a task-specific start token \textspt{[S]}. At each step, beam search simply selects top-$k$ next tokens with the largest attention scores, and just keeps $k$ partial candidates with the highest scores, where $k$ is the beam size. When a candidate produces a task-specific end token \textspt{[E]}, the candidate is complete. For OIE, \textspt{[S]} and \textspt{[E]} refer to the argument pair, e.g, (\textspt{[S]} is ``Fisher'', and \textspt{[E]} refers to ``London Opera Centre''). 

The above traditional beam search only allows searching sequences between \textspt{[S]} and \textspt{[E]}. To adapt beam search to produce more triples, we allow searching sequences: (\expandafter{\romannumeral1}) left to both \textspt{[S]} and \textspt{[E]}, and (\expandafter{\romannumeral2}) right to both \textspt{[S]} and \textspt{[E]}. This helps to improve the recall of the candidates. For example, a candidate triple (Fisher; Born in; Glasgow) is generated by looking at ``Born in'' on the left in the above example. We also need to enable bidirectionality by running the search in both directions (left to right and right to left) following \citet{wang2020language}. For OIE, we implement this by allowing every argument as both \textspt{[S]} and \textspt{[E]} regardless of its position in the input. For example, ``Fisher'' is \textspt{[S]} in (Fisher; Born in; Glasgow) given ``Glasgow'' appears before ``Fisher'' in the input.

\paragraph{Ranking} The ranking stage finds triples that are of interest to the task via a ranking model pre-trained on a task-agnostic relational corpus. For OIE, the generating stage produces $k$ candidate triples for every argument pair. However, the sequences in the candidates are relevant to the argument pairs, not just in the relational aspect. The ranking stage aims to find the triples that specifically express the relational information between the argument pair, which is important for OIE.

We propose a contrastive model to rank the triples as illustrated in Figure~\ref{fig:oie}. Given a batch of $N$ (sentence, triple) pairs, the model is trained to predict which of the $N^2$ possible (sentence, triple) pairs across a batch actually appeared. The model learns a joint embedding space by training a base encoder BERT. The input sequence of the BERT encoder is in the format: \textspt{[CLS]} sentence \textspt{[SEP]} triple \textspt{[SEP]}, which follows the standard input format of BERT. The goal is to maximize the cosine similarity of the sentence and triple embeddings of the $N$ true pairs in the batch while minimizing the cosine similarity of the embeddings of the remaining $N^2-N$ incorrect pairs. We optimize a cross-entropy loss over these similarity scores. The loss function for a positive pair is defined by $l$ in Eq.~\ref{eq:cf}.

{\small
\begin{equation*}
    l_{\rm{sentence}} = - \log \frac{\exp(\rm{sim}(\mathbf{u}_i, \mathbf{v}_i))}{\sum_{k=1}^{N} \exp(\rm{sim}(\mathbf{u}_i, \mathbf{v}_k))} \\ \nonumber
\end{equation*}
\begin{equation*}
    l_{\rm{triple}} = - \log \frac{\exp(\rm{sim}(\mathbf{u}_i, \mathbf{v}_i))}{\sum_{k=1}^{N} \exp(\rm{sim}(\mathbf{u}_k, \mathbf{v}_i))}
\end{equation*}
\begin{equation}
    l = \frac{l_{\rm{sentence}}+l_{\rm{triple}}}{2}
    \label{eq:cf}
\end{equation}
}
where sim$(\mathbf{u}, \mathbf{v}) = \frac{\mathbf{u}^\intercal \mathbf{v}}{\Vert \mathbf{u} \Vert \Vert \mathbf{v} \Vert}$. For the $i$-th positive (sentence, triple) pair, $\mathbf{u}_i$ and $\mathbf{v}_i$ denote the sentence and triple embedding respectively. 

We take advantage of the pre-trained BERT$_{\rm BASE}$ as the base encoder. We further simplify the standard contrastive learning framework by removing the projection layer between the representation and the contrastive embedding space. Neither the linear~\cite{radford2021learning} nor non-linear~\cite{chen2020simple} projection is used. This is because sentences and triples are unified in the same embedding space of BERT. We train the model on T-REx~\cite{elsahar2019t}, which is a dataset of large-scale alignments between Wikipedia abstracts and Wikidata triples. T-REx contains a large number of sentence-triple pairs (11 million triples are paired with 6.2 million sentences). T-REx also reports an accuracy of 97.8\% of the pairs. 

The ranking model is task-agnostic. The ranking model takes the input in the same format for all tasks. At test time, the input text and each candidate triple from the generating stage is paired as the input to the ranking model. The candidate triples are ranked by the contrastive loss. We adopt the top-$n$ candidate triples returned by the ranking model as the output. $n$ varies across different tasks~\footnote{\small{Please refer to Appendix~\ref{sec:expsetup} for details.}}. For the above OIE example, the output is the top-2 triples.

\subsection{Decoding}
Once the output triples are produced, we decode the output triples to obtain task predictions. For OIE, the output triples serve as task predictions directly. No specific decoding strategy is needed.

\section{Information Extraction Tasks}
\subsection{Open Information Extraction} 
The details are provided in Sec.~\ref{sec:method}.

\subsection{Relation Classification} 
\label{sec:rc}
For this task, we are given an input sentence with gold head and tail entities aiming to classify the relation type in a pre-defined category. 

\paragraph{Input and Output Format} The input is a sentence encoded with gold head and tail entities, and linked relation phrases. The output is a triple. An example is below.
\begin{enumerate*}
    \item[] {\bf Input} \underline{Born in}$_{\rm place\_of\_birth}$ \underline{Glasgow}$_{\rm GOLD}$, \underline{Fisher}$_{\rm GOLD}$ is a graduate of the London Opera Centre.
    \item[] {\bf Output} (Fisher; place\_of\_birth; Glasgow).
\end{enumerate*}
${\rm GOLD}$ denotes the gold entity. The linked relation phrases annotated with Wikidata predicates, e.g., \underline{Born in}$_{\rm place\_of\_birth}$, are constructed as follows. We use an offline dictionary that maps the pre-defined relations to the Wikidata predicates. Such dictionaries are often provided either by Wikidata or third-parties. In all tested datasets, we can refer to either gold Wikidata or other high-quality resources for the dictionaries. We consider a sequence of tokens linked to a certain relation if the tokens are matched with the label or alias of the particular predicate in Wikidata. In the above example, ``Born in'' matches an alias of the Wikidata predicate ``place\_of\_birth''. In practice, some Wikidata predicates do not provide as many aliases as others. Inspired by \citet{angeli2015leveraging}, we follow the below procedure to add new aliases to resolve the imbalance issue: We first create a large candidate set of Wikipedia relations aligned to Wikidata predicates via distant supervision, then ask human annotators to filter out the wrong alignments. The remaining aligned relation phrases are added as new aliases of the Wikidata predicates.

\paragraph{Relation-Constrained Translation} For the beam search in the generating stage of Sec.~\ref{sec:gen}, \textspt{[S]} and \textspt{[E]} are the gold head and tail entities respectively. As the task requires the relations to be from a pre-defined category, using the beam search directly is not efficient. Allowing generating any token at each step might lead to sequences that do not match any pre-defined relations. Similar to \citet{de2020autoregressive}, we use constrained beam search, which only decodes tokens belonging to a linked relation phrase. We take the top-$n$ triples from the ranking model as the output.

\paragraph{Decoding Relation} We take the Wikidata predicates of the output triples, and map the predicates back to the relations in the pre-defined category, which serve as the task predictions. In the above input/output example, ``place\_of\_birth'' is the Wikidata predicate in the output triple. It is mapped to ``city\_of\_birth'' in the pre-defined relation category of one of the widely used relation classification datasets, TACRED. ``city\_of\_birth'' hence serves as the task prediction.

\subsection{Factual Probe}
\label{sec:fr}
Given an input sentence with gold head entity name and relation name, the task aims to fill in the tail entity.

\paragraph{Input and Output Format} The input is encoded as a NP-chunked sentence with gold head entity candidates and linked relation phrases. The output is a triple. An example is below.
\begin{enumerate*}
    \item[] {\bf Input} \underline{Born in}$_{\rm place\_of\_birth}$ \underline{Glasgow}$_{\rm NP}$, \underline{Fisher}$_{\rm GOLD/NP}$ is a graduate of the \underline{London Opera Centre}$_{\rm NP}$.
    \item[] {\bf Output} (Fisher; place\_of\_birth; Glasgow).
\end{enumerate*}
${\rm GOLD/NP}$ denotes the noun phrase that matches the gold head entity. \underline{Born in}$_{\rm place\_of\_birth}$ represents a linked relation phrase annotated with a Wikidata predicate which is constructed in the same way as in Sec.~\ref{sec:rc}.

\paragraph{Entity-Constrained Translation} For the beam search, \textspt{[S]} and \textspt{[E]} are the gold head entity candidate and linked relation phrase respectively. Similar to the relation classification, we also constrain the search to generate possible tail entity sequences. We assume that NPs other than the gold head entity provide the set of candidate tail entities. To enable this, the search only decodes tokens belonging to the candidate NPs. In practice, we take the top-1 triple from the ranking model as the output.

\paragraph{Decoding Tail Entity} We take the tail entities of the output triples as task predictions. For example, in the above output triple, ``Glasgow'' is decoded as the task prediction.

\section{Experiments}
In this section, we show that \method\ framework solves the different information extraction tasks considered and outperforms the task-specific state-of-the-art results on multiple datasets.

To keep the framework as simple as possible, most settings and hyperparameters are shared across all experiments. For example, we use BERT$_{\rm LARGE}$~\cite{devlin2018bert} for the beam search of the generating stage (Sec.~\ref{sec:gen}) for all tasks. The details of the experimental setup, datasets, and comparison methods are described in Appendix~\ref{sec:expsetup}.

\subsection{Main Results}
\begin{figure*}
\centering
\subcaptionbox{{\small OIE2016.}}{\includegraphics[width=0.25\textwidth]{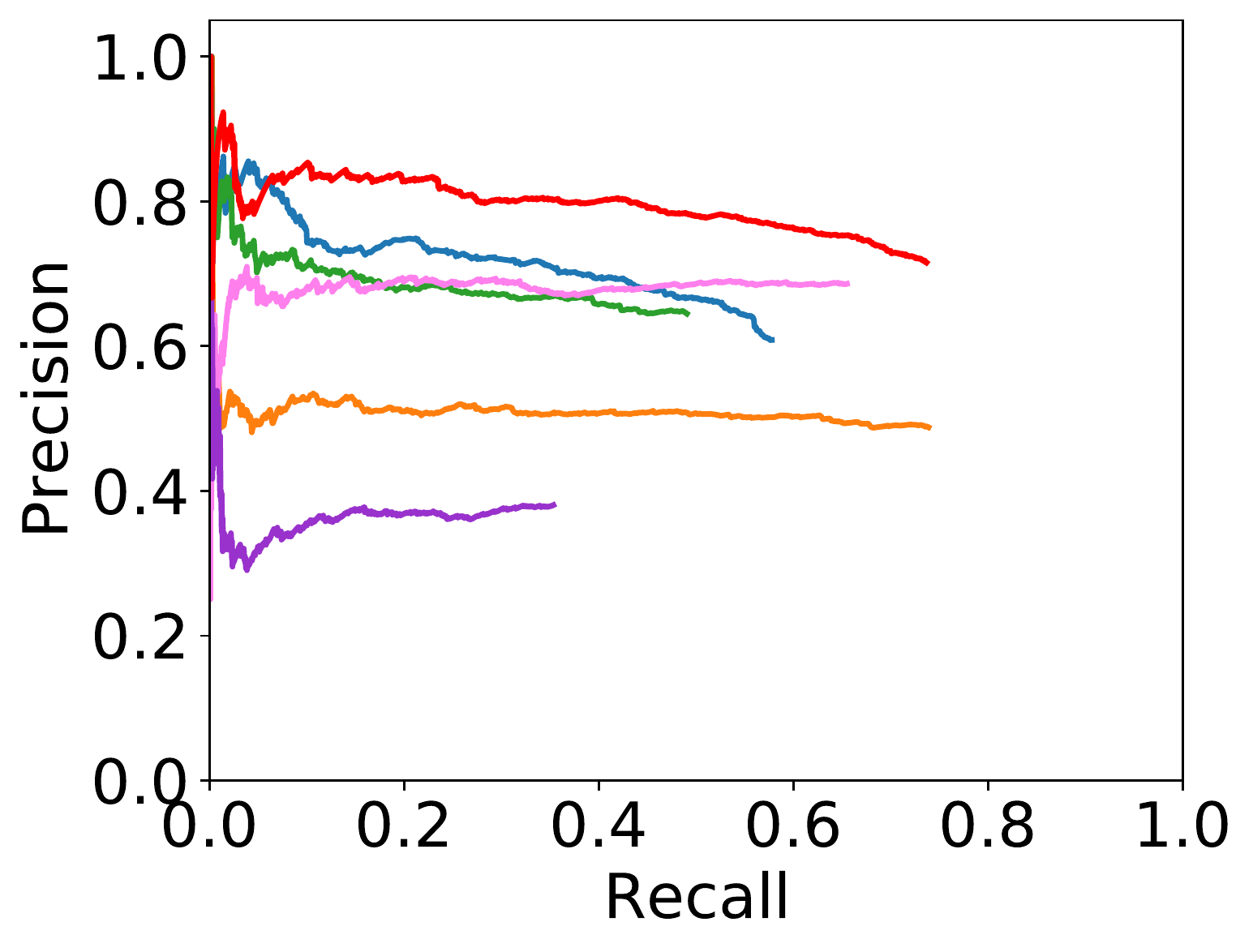}}%
\subcaptionbox{{\small WEB.}}{\includegraphics[width=0.25\textwidth]{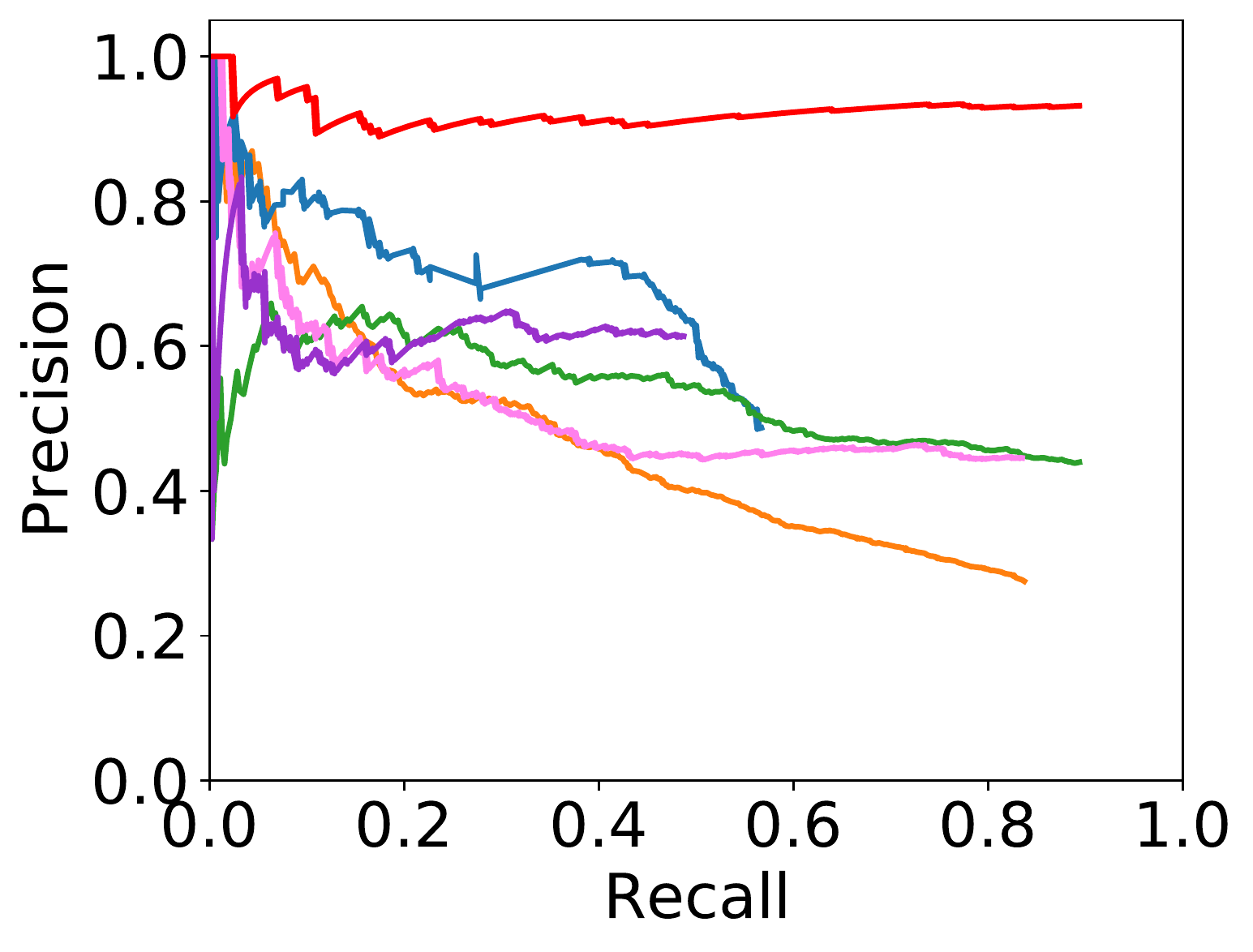}}%
\subcaptionbox{{\small NYT.}}{\includegraphics[width=0.25\textwidth]{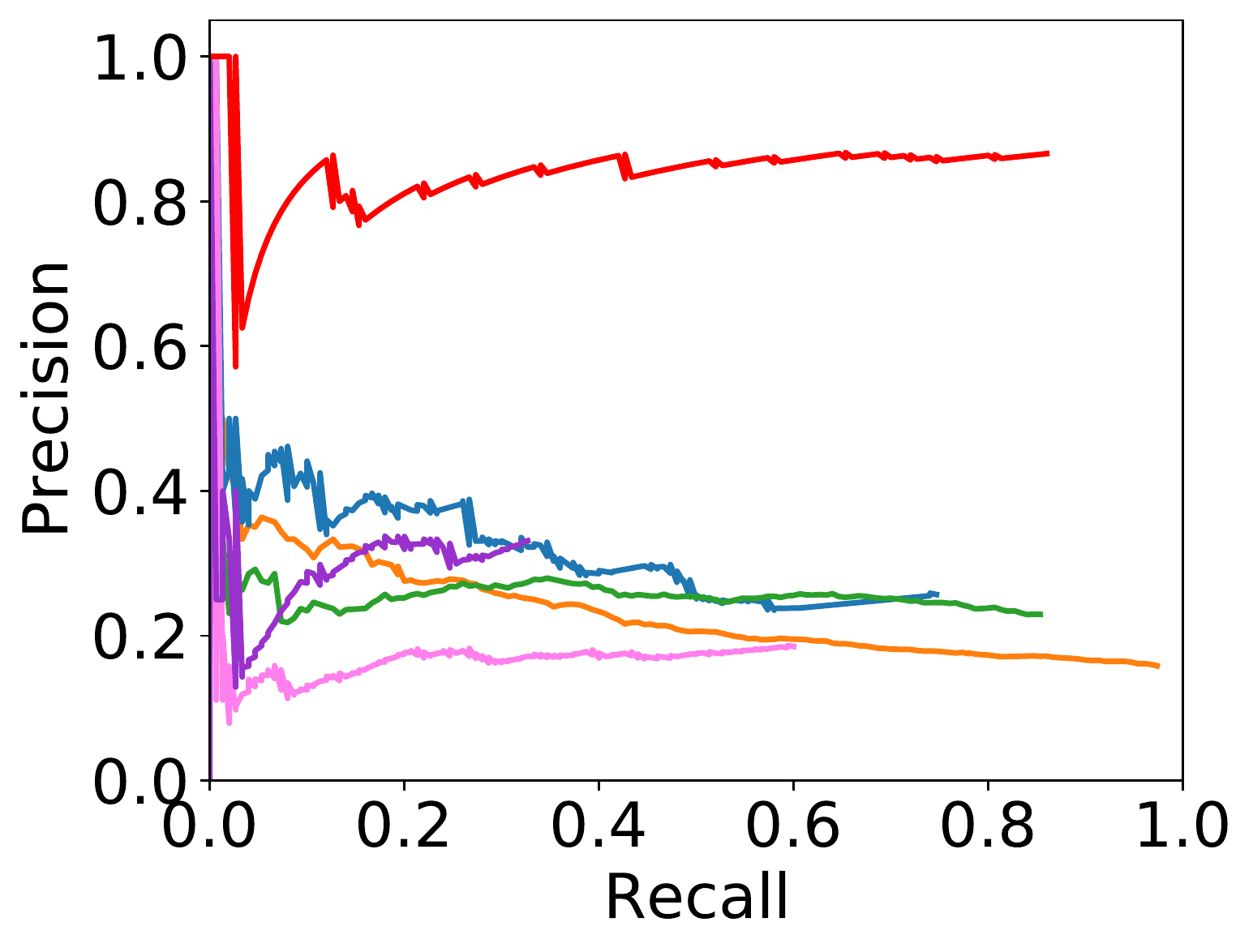}}%
\subcaptionbox{{\small PENN.}}{\includegraphics[width=0.25\textwidth]{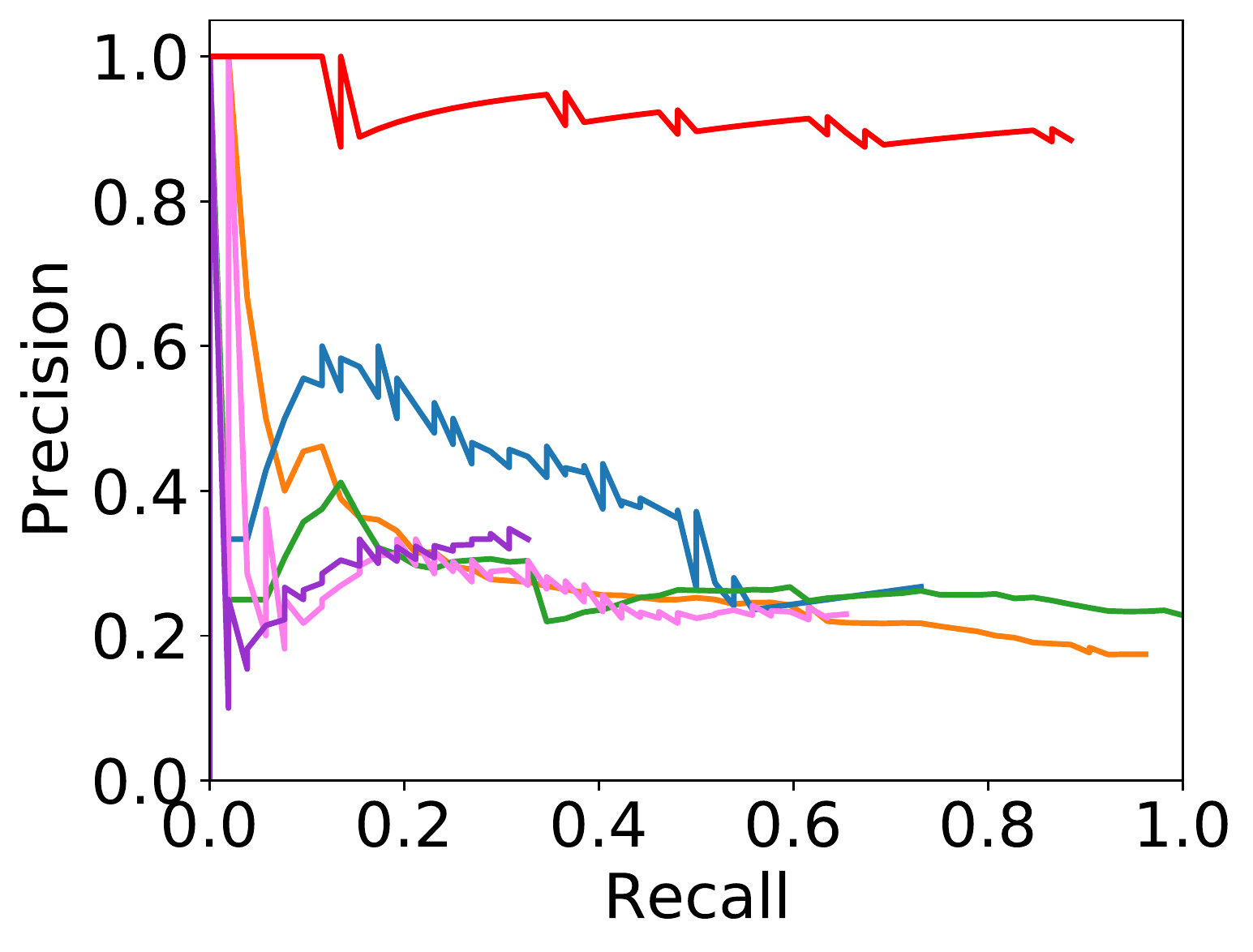}}%
\\
\includegraphics[width=0.7\linewidth]{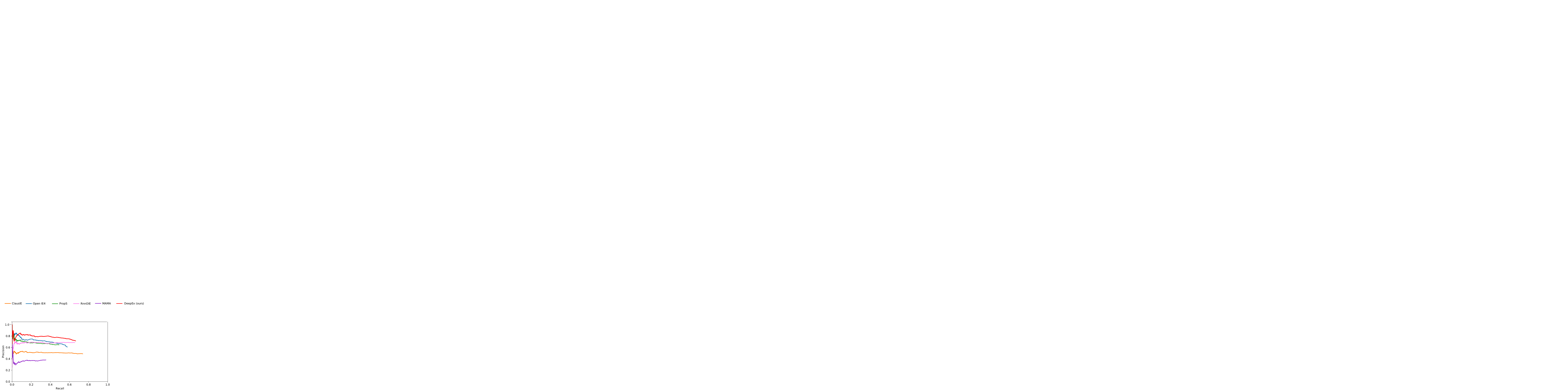}%
\caption{{\small Precision-recall curves of the different open information extraction (OIE) systems on OIE datasets.\zui{add}}}
\label{fig:prc}
\end{figure*}

The results are shown in Table~\ref{tab:allres}. We achieve state-of-the-art results on the following datasets in a zero-shot setting even outperforming fully supervised methods: (\expandafter{\romannumeral1}) Open information extraction (OIE): OIE2016, WEB, NYT, PENN; and (\expandafter{\romannumeral2}) Factual probe: T-REx. The improvements are significant for OIE. In particular, the zero-shot \method\ outperforms RnnOIE by on average 37.5 in F1 and 44.6 in AUC, which is a supervised OIE system introduced in \cite{stanovsky2018supervised}. Given no specific OIE training data is used by \method, the results are encouraging, suggesting that the zero-shot transfer of the latent knowledge that a pre-trained LM has about the tasks is successful. State-of-the-art OIE performance is obtained without referring to task-specific training data, and such zero-shot ability is generalizable across multiple datasets. The PR curves for all OIE test sets are depicted in Figure~\ref{fig:prc}. Similar to the findings in Table~\ref{tab:allres}, overall, \method\ outperforms the comparison systems across all datasets. For each dataset, it provides a superior precision-recall curve. \method\ slightly outperforms the comparison methods on T-REx (factual probe). The main reason is that the task-specific LAMA~\cite{petroni2020context} can use the wrong memory of LMs to answer without needing the mention of the tail entity. An example expressing the triple (Nicholas Liverpool; place\_of\_death; Miami) is shown in Table~\ref{tab:lamamistake} in Appendix. Thanks to the explainability and transparency of our framework, we can avoid such errors. The results demonstrate that the zero-shot \method\ generalizes well to different information extraction tasks.  
\begin{table*}[]
\centering
\resizebox{0.9\textwidth}{!}{%
\begin{tabularx}{\textwidth}{l*{9}{C}}
\multicolumn{9}{c}{\textbf{Open Information Extraction} (F1 and AUC)}  \\ \midrule
 & \multicolumn{2}{c}{\textbf{OIE2016}} & \multicolumn{2}{c}{\textbf{WEB}} & \multicolumn{2}{c}{\textbf{NYT}} & \multicolumn{2}{c}{\textbf{PENN}} \\ \cmidrule(l){2-9} 
ClausIE~\cite{del2013clausie}  & 58.8   & 37.6  & 44.9 & 40.1& 29.6 & 22.9& 34.6 & \secbest 28.4  \\
Open IE4~\footref{ft:openie51}& 59.6   & 41.7  & 55.7 & 40.5& \secbest 38.3 & \secbest 24.0& \secbest 42.6 & 28.1 \\
PropS~\cite{stanovsky2016getting}  & 55.6   & 33.8  & \secbest 58.9 & \secbest 48.0& 37.2 & 22.1& 39.1 & 27.7 \\
RnnOIE$^*$~\cite{stanovsky2018supervised}  & \secbest 67.0   & \secbest 44.5  & 58.1 & 43.5& 28.3 & 10.4& 34.5 & 18.3 \\
MAMA~\cite{wang2020language}& 36.6  & 12.8 & 54.3  & 30.3   & 32.9& 9.4   & 33.0& 9.0 \\
\addlinespace
\textbf{\method} (Zero-Shot) (\electricblue{\small ours}) & \best 72.6  & \best 58.6 & \best 91.2 & \best 82.4& \best 85.5 & \best 72.5& \best 88.5& \best 81.5 
\end{tabularx}}\par\vskip+16pt
\resizebox{0.85\textwidth}{!}{%
\begin{tabularx}{\textwidth}{l*{6}{C}}
\multicolumn{6}{c}{\textbf{Relation Classification} (F1)}  \\ \midrule
\multirow{3}{*}{\textbf{}} & \multirow{3}{*}{\textbf{TACRED}} & \multicolumn{4}{c}{\textbf{FewRel 1.0 (dev)}}  \\ \cmidrule(l){3-6} 
  &  & 5-way  & 5-way  & 10-way & 10-way \\
  &  & 1-shot & 5-shot & 1-shot & 5-shot \\ \cmidrule(l){2-6} 
BERT-EM$^*$~\cite{soares2019matching}   & 70.1 & 88.9   &-& 82.8   & - \\
BERT$_{\rm EM}$+MTB$^*$~\cite{soares2019matching}& 71.5 & 90.1   &   - & \secbest 83.4   & - \\
DG-SpanBERT$^*$~\cite{chen2020efficient}& 71.5 &  -  &   - &  -  & - \\
BERT-PAIR$^*$~\cite{gao2019fewrel}  &-  & 85.7   & 89.5   & 76.8   & 81.8 \\
TANL$^*$~\cite{paolini2021structured}   & \secbest 71.9 & \best 94.0   & \best 96.4   & 82.6   & \secbest 88.2  \\ \addlinespace
\textbf{\method} (Zero-Shot Top-1) (\electricblue{\small ours})   & 49.2  &48.8&   48.8&48.8   &   48.8   \\
\textbf{\method} (Zero-Shot Top-10) (\electricblue{\small ours}) &   \best 76.4   &\secbest 92.9&\secbest 92.9& \best 92.9  & \best 92.9  
\end{tabularx}}\par\vskip+16pt
\resizebox{0.85\textwidth}{!}{%
\begin{tabularx}{\textwidth}{l*{6}{C}}
\multicolumn{6}{c}{\textbf{Factual Probe} (P@1)}  \\ \hline
\textbf{}& \multicolumn{4}{c}{\textbf{Google-RE}} & \textbf{T-REx} \\ \cline{2-6} 
 & birth-place & birth-date & death-place & Total & Total  \\ \cline{2-6} 
LAMA-Original~\cite{petroni2019language}& 16.1& 1.4& 14.0& 10.5  & 32.3   \\
LAMA-Oracle~\cite{petroni2020context}  & \best 70.6& \best 98.1   & \best 65.1& \best 78.0  & \secbest 62.6   \\ \addlinespace
\textbf{\method} (Zero-Shot) (\electricblue{\small ours}) &   \secbest 67.8  & \secbest 91.0   &\secbest 64.1 &  \secbest 74.3 & \best 66.0  
\end{tabularx}}%
\caption{{\small Results on all tasks. All evaluation scores are higher the better. An asterisk (*) indicates a supervised method.  \zui{add}\haoyun{add}\xiao{add}}}
\label{tab:allres}
\end{table*}

\begin{table*}[]
\resizebox{\linewidth}{!}{
\begin{tabular}{@{}lp{16cm}lll@{}}
\toprule
{\bf Dataset}                    & {\bf Example (sentence and gold triple)}                                                                                                                           & {\bf LAMA-Oracle}         & {\bf \method} (\electricblue{\small ours}) & {\bf Error Type}            \\ \midrule
\multirow{4}{*}{Google-RE} & {\sl Benny Marinelli (c. {\bf 1902} -- October 22, 1927) was an American Thoroughbred horse racing jockey best known for winning the Classic Preakness Stakes in 1923 .} & \multirow{2}{*}{1902 \cmark} & \multirow{2}{*}{- \xmark} & \multirow{2}{*}{Missing relation} \\
                           & (Benny Marinelli; birth\_date; {\bf 1902})                                                                                                                          &                       &                    \\ \cmidrule(l){2-5} 
                           & {\sl Myer Hoffman (21 July 1902 in {\bf Leeds}, Yorkshire, England – 14 October 1959 in Lourenço Marques, Mozambique) was an English-born Irish cricketer.} & \multirow{2}{*}{Leeds \cmark}     & \multirow{2}{*}{- \xmark}  & \multirow{2}{*}{Missing relation}\\
                           & (Myer Hoffman; place\_of\_birth; {\bf Leeds})                                                                                                                   &                       &                    \\ \midrule
\multirow{4}{*}{T-REx}     & {\sl ..., and Bulman, as priest Frank Kane in {\bf BBC} drama The Paradise Club (1989–90), and as General Tagge in the first Star Wars film (1977).} & \multirow{2}{*}{BBC \cmark}     & \multirow{2}{*}{- \xmark}  & \multirow{2}{*}{Missing relation}\\
                           & (The Paradise Club; original\_network; {\bf BBC})                                                                                                                &                       &                    \\ \cmidrule(l){2-5}
                           & {\sl Judges` lodgings, the house once occupied by former Prime Minister Edward Heath at {\bf Salisbury}.} & \multirow{2}{*}{Salisbury \cmark}     & \multirow{2}{*}{- \xmark}  & \multirow{2}{*}{Wrong relation}\\
                           & (Edward Heath; place\_of\_death; {\bf Salisbury})                                                                                                             &                       &                    \\
                           \bottomrule
\end{tabular}}
\caption{{\small Error analysis of \method\ on factual probing datasets. \xiao{add}}}
\label{tab:lamaerror}
\end{table*}

For other datasets, we obtain comparable performance with the best comparison methods. We highlight that our approach uses a unified framework that tackles all the tasks in a zero-shot way. Our framework is task-agnostic without task-specific training or module modification, which is in contrast with task-specific models trained on specific corpora as shown in Table~\ref{tab:allres}. For relation classification, all the comparison methods are fully supervised and trained on the corresponding large-scale corpora. Our top-1 zero-shot result serves as a lower bound, while top-10 results indicate an ideal situation when an improved ranking model is available. Interestingly, on FewRel, a benchmark for few-shot relation classification, our top-10 zero-shot performance sometimes is the best. While TACRED is not specifically a few-shot dataset, there are many label types that rarely appear in the training set~\cite{paolini2021structured}. This shows the importance of zero-shot information extraction in low-resource regimes. The ranking model is based on BERT$_{\rm BASE}$. It is interesting to check whether larger pre-trained LMs (e.g., BERT$_{\rm LARGE}$) are more capable of ranking. We plan to investigate this in the future. On the other factual probing dataset, Google-RE, we perform slightly worse compared to LAMAs. This is mainly due to the missing mentions of relations in the sentences as shown in Table~\ref{tab:lamaerror}. 

\subsection{Error Analysis}
\label{sec:err}
To better understand the limitations of \method, we perform a detailed analysis of errors in its recall as \method\ lacks more in recall compared to precision across all the datasets. We use open information extraction as an example. We only show F1 and AUC in Table~\ref{tab:allres}, and Figure~\ref{fig:prc} illustrates the precision-recall curves showing recall errors are the main limitation. We therefore randomly sample 50 recall errors made by \method\ on the WEB corpus and summarized the types of common errors as below. We find 46\% of the errors are due to the spaCy noun chunker identifying the wrong arguments. 12\% of the recall errors are cases where the predicate is a noun or nominalized. 10\% of the examined errors are involved in long sentences. Details are described in Table~\ref{tab:errmain} in Appendix.

While most of the error types are shared across the datasets, we find a type of error due to the explainability and transparency of \method, which we cannot avoid. The error is mainly due to the missing mention of relations in the sentences. This type of error mainly appears in factual probing and relation classification datasets. The reason is that the tasks do not require the existence of the actual relation span in the input. The tasks often provide the relation as an input or the relation is expressed in a vague way that can not be linked to a predicate. We take factual probe as an example in Table~\ref{tab:lamaerror}. A sentence is given to express the ``place\_of\_death'' relation can only contain mentions of ``residence'' relation such as ``occupied by''. While the gold data might consider this as a correct prediction, \method\ uses triples extracted from the sentences. We sacrifice performance for better explainability and transparency. We believe it is ideal to allow a trade-off between performance and explainability. We leave this as future work. Also, ``birth date'' can be expressed as ``(c.''. Again in such cases, we sacrifice performance for explainability.

\begin{table}[]
\centering
\resizebox{0.9\linewidth}{!}{
\begin{tabular}{@{}lllll@{}}
\toprule
{\bf Method}        & {\bf P} & {\bf R} & {\bf F1} & {\bf AUC} \\ \midrule
{\bf \method} (\electricblue{\small ours})                & 70.9 & 73.8 & 72.3 & 57.4    \\
No beam search-RnnOIE & 72.2 & 60.8 & 66.0 & 49.9 \\
\hspace{0.1in} RnnOIE & 70.0 & 66.8 & 68.3 & 47.1 \\
In-between beam search & 68.9 & 65.7 & 67.3 & 50.3 \\
Beam search-BERT$_{\rm BASE}$      & 71.9 & 66.7 & 69.2 & 53.0 \\
No ranking model       & 39.6 & 38.1 & 38.8 & 13.8 \\ \bottomrule
\end{tabular}
}
\caption{{\small Ablation over different facets of \method\ on OIE2016 dev set.}}
\label{tab:ablationmain}
\end{table}

\subsection{Ablation Studies}
We perform ablation experiments to understand the relative importance of different facets of \method. We demonstrate the importance of the generating stage in particular beam search and the size of the pre-trained LM, and the ranking stage in particular the ranking model on open information extraction. We evaluate the below settings on the OIE2016 dev set. The first three settings examine the relative importance of the beam search: (\expandafter{\romannumeral1}) No beam search-RnnOIE: Instead of using the beam search, we use the best supervised OIE system RnnOIE in the generating stage. All the other components of \method\, including the ranking model, are kept the same. We add the results of RnnOIE as well for comparison. (\expandafter{\romannumeral2}) In-between beam search: We only allow searching sequences between the argument pair while keeping the other components of \method\ the same as the default. (\expandafter{\romannumeral3}) Beam search-BERT$_{\rm BASE}$: Instead of BERT$_{\rm LARGE}$, we use BERT$_{\rm BASE}$ for beam search. (\expandafter{\romannumeral4}) No ranking model: We do not leverage the ranking model, and directly rank the candidates using their original scores based on the attention from the generating stage.

We first examine the effect brought by the beam search. As shown in Table~\ref{tab:ablationmain}, we find removing beam search of the generating stage greatly hurts the performance. \method\ outperforms the best supervised OIE system by 6.3 in F1 and 7.5 in AUC. The result confirms our intuition that pre-trained LMs enable the zero-shot transfer of the latent knowledge that they have about the task. The original RnnOIE performs similarly; this is due to the training set of RnnOIE which provides good coverage of the triples on the dev set. We secondly study the importance of the triple-oriented beam search. We find limiting the search significantly hurts the performance. It is often that the triples are expressed in inverted sentences, such as (Fisher; Born in; Glasgow) from ``Born in Glasgow, Fisher is a graduate of the London Opera Centre''. In fact, a considerable amount of gold triples containing valid relation sequences appear outside the argument pair. For example, 16.9\% of the relation sequences are not between the argument pairs on the OIE2016 test set. More results are shown in Appendix~\ref{sec:aar}. We then test the impact of the size of the pre-trained LM. We find that BERT$_{\rm BASE}$ performs worse than BERT$_{\rm LARGE}$. This indicates that larger pre-trained LMs (e.g., BERT$_{\rm LARGE}$) provide more general knowledge about the task that improves the results. Next, we study the impact of the ranking model. We find that removing the ranking model significantly hurts the performance. The results suggest that the ranking model can distinguish the relational triples from the rest among the candidates.

\section{Related Work}
Relation classification aims to identify the correct relation type from a pre-defined set of relations between two given entities. Language models (LM)~\cite{xu2021pre} pre-trained with self-supervised~\cite{liu2021self} objectives, e.g., BERT~\cite{devlin2018bert}, GPT~\cite{radford2018improving,radford2019language,brown2020language}, RoBERTa~\cite{abs-1907-11692}, transfer well to relation classification datasets in fine-tuning~\cite{joshi2020spanbert,gao2019fewrel} or few-shot regime~\cite{soares2019matching} with architecture modifications. Sequence-to-sequence models, such as T5~\cite{raffel2019exploring}, BART~\cite{lewis2019bart} and GLM~\cite{du2021all}, are adapted to the task based on data augmentation and fine-tuning~\cite{paolini2021structured}. Besides relation classification, \citet{paolini2021structured} generalize T5 to some more structured prediction tasks as well, e.g., semantic role labeling and event extraction. However, \method\ enables zero-shot relation classification that does not require any task-specific training.

Many open information extraction (OIE) systems, e.g., Stanford OpenIE~\cite{angeli2015leveraging}, OLLIE~\cite{schmitz2012open}, Reverb~\cite{fader2011identifying}, and their descendant Open IE4 leverage carefully-designed linguistic patterns (e.g., based on dependencies and POS tags) to extract triples from textual corpora without using additional training sets. Recently, supervised OIE systems~\cite{stanovsky2018supervised,ro2020multi,kolluru2020oie6} formulate the OIE as a sequence generation problem using neural networks trained on additional training sets. Similar to our work, \citet{wang2020language} use the parameters of LMs to extract triples, with the main difference that \method\ not only improves the recall of the beam search, but also uses a pre-trained ranking model to enhance the zero-shot capability.

LMs are used in factual probing tasks, by using the outputs alone~\cite{petroni2019language} to answer the relation-specific queries in cloze statements. \citet{petroni2020context} additionally feed sentences expressing the facts to the LMs and shows improved results. Other than template-based queries, learning trigger-based~\cite{shin2020autoprompt} and continuous prompts~\cite{liu2021gpt,li2021prefix} are helpful in recalling the facts. The main difference is that \method\ explores the internal parameters of the LMs rather than the outputs, and the results are more interpretable.

Overall, in contrast to the existing approaches, \method\ unifies the open information extraction, relation classification, and factual probe under the same framework in zero-shot settings.
\section{Conclusion}
We have demonstrated that our unified approach can handle multiple information extraction tasks within a simple framework and shows improvements in zero-shot settings. Unlike previous approaches designing complicated task-specific pipelines, \method\ enables conducting all considered information extraction tasks with only input and output design. Therefore, \method\ is flexible and can be adapted to a variety of tasks. Different from previous approaches that target pre-defined categories (e.g., fixed relation types for relation classification), \method\ generalizes better to unseen classes as the generating stage leverages the transfer of latent knowledge that a pre-trained language model has about the tasks. Besides, the ranking stage pre-trains on a large-scale task-agnostic dataset. \method\ exhibits strong zero-shot capabilities in low-resource tasks without the need of any task-specific training set. \method\ also exploits the in-depth information of the language models, i.e., parameters, rather than the outputs alone, which enhances the explainability through enhanced model transparency. Based on our findings, we believe that the unified approach advances the research in understanding natural language semantics (e.g., structure prediction tasks) using deep learning models. We hope our results will foster further research in this direction.
\section*{Acknowledgement}
We would like to thank the anonymous reviewers for their suggestions and comments. This material is in part based upon work supported by Berkeley DeepDrive and Berkeley Artificial Intelligence Research.

\bibliography{anthology,custom}

\begin{thebibliography}{42}
\expandafter\ifx\csname natexlab\endcsname\relax\def\natexlab#1{#1}\fi

\bibitem[{Angeli et~al.(2015)Angeli, Premkumar, and
  Manning}]{angeli2015leveraging}
Gabor Angeli, Melvin Jose~Johnson Premkumar, and Christopher~D Manning. 2015.
\newblock Leveraging linguistic structure for open domain information
  extraction.
\newblock In \emph{ACL}, pages 344--354.

\bibitem[{Bhardwaj et~al.(2019)Bhardwaj, Aggarwal, and
  Mausam}]{bhardwaj2019carb}
Sangnie Bhardwaj, Samarth Aggarwal, and Mausam Mausam. 2019.
\newblock Carb: A crowdsourced benchmark for open ie.
\newblock In \emph{EMNLP}, pages 6263--6268.

\bibitem[{Brown et~al.(2020)Brown, Mann, Ryder, Subbiah, Kaplan, Dhariwal,
  Neelakantan, Shyam, Sastry, Askell et~al.}]{brown2020language}
Tom~B Brown, Benjamin Mann, Nick Ryder, Melanie Subbiah, Jared Kaplan, Prafulla
  Dhariwal, Arvind Neelakantan, Pranav Shyam, Girish Sastry, Amanda Askell,
  et~al. 2020.
\newblock Language models are few-shot learners.
\newblock In \emph{NeurIPS}, volume~33, pages 1877--1901.

\bibitem[{Chen et~al.(2020{\natexlab{a}})Chen, Hoehndorf, Elhoseiny, and
  Zhang}]{chen2020efficient}
Jun Chen, Robert Hoehndorf, Mohamed Elhoseiny, and Xiangliang Zhang.
  2020{\natexlab{a}}.
\newblock Efficient long-distance relation extraction with dg-spanbert.
\newblock \emph{arXiv preprint arXiv:2004.03636}.

\bibitem[{Chen et~al.(2020{\natexlab{b}})Chen, Kornblith, Norouzi, and
  Hinton}]{chen2020simple}
Ting Chen, Simon Kornblith, Mohammad Norouzi, and Geoffrey Hinton.
  2020{\natexlab{b}}.
\newblock A simple framework for contrastive learning of visual
  representations.
\newblock In \emph{ICML}, pages 1597--1607. PMLR.

\bibitem[{De~Cao et~al.(2021)De~Cao, Izacard, Riedel, and
  Petroni}]{de2020autoregressive}
Nicola De~Cao, Gautier Izacard, Sebastian Riedel, and Fabio Petroni. 2021.
\newblock Autoregressive entity retrieval.
\newblock In \emph{ICLR}.

\bibitem[{Del~Corro and Gemulla(2013)}]{del2013clausie}
Luciano Del~Corro and Rainer Gemulla. 2013.
\newblock Clausie: clause-based open information extraction.
\newblock In \emph{WWW}, pages 355--366.

\bibitem[{Devlin et~al.(2019)Devlin, Chang, Lee, and
  Toutanova}]{devlin2018bert}
Jacob Devlin, Ming-Wei Chang, Kenton Lee, and Kristina Toutanova. 2019.
\newblock Bert: Pre-training of deep bidirectional transformers for language
  understanding.
\newblock In \emph{NAACL}, pages 4171--4186.

\bibitem[{Du et~al.(2021)Du, Qian, Liu, Ding, Qiu, Yang, and Tang}]{du2021all}
Zhengxiao Du, Yujie Qian, Xiao Liu, Ming Ding, Jiezhong Qiu, Zhilin Yang, and
  Jie Tang. 2021.
\newblock All nlp tasks are generation tasks: A general pretraining framework.
\newblock \emph{arXiv preprint arXiv:2103.10360}.

\bibitem[{Elsahar et~al.(2019)Elsahar, Vougiouklis, Remaci, Gravier, Hare,
  Simperl, and Laforest}]{elsahar2019t}
Hady Elsahar, Pavlos Vougiouklis, Arslen Remaci, Christophe Gravier, Jonathon
  Hare, Elena Simperl, and Frederique Laforest. 2019.
\newblock T-rex: A large scale alignment of natural language with knowledge
  base triples.

\bibitem[{Fader et~al.(2011)Fader, Soderland, and
  Etzioni}]{fader2011identifying}
Anthony Fader, Stephen Soderland, and Oren Etzioni. 2011.
\newblock Identifying relations for open information extraction.
\newblock In \emph{EMNLP}, pages 1535--1545.

\bibitem[{Gao et~al.(2019)Gao, Han, Zhu, Liu, Li, Sun, and
  Zhou}]{gao2019fewrel}
Tianyu Gao, Xu~Han, Hao Zhu, Zhiyuan Liu, Peng Li, Maosong Sun, and Jie Zhou.
  2019.
\newblock Fewrel 2.0: Towards more challenging few-shot relation
  classification.
\newblock In \emph{EMNLP}, pages 6250--6255.

\bibitem[{Han et~al.(2021)Han, Zhang, Ding, Gu, Liu, Huo, Qiu, Zhang, Han,
  Huang et~al.}]{xu2021pre}
Xu~Han, Zhengyan Zhang, Ning Ding, Yuxian Gu, Xiao Liu, Yuqi Huo, Jiezhong Qiu,
  Liang Zhang, Wentao Han, Minlie Huang, et~al. 2021.
\newblock Pre-trained models: Past, present and future.
\newblock \emph{AI Open}.

\bibitem[{Han et~al.(2018)Han, Zhu, Yu, Wang, Yao, Liu, and
  Sun}]{han2018fewrel}
Xu~Han, Hao Zhu, Pengfei Yu, Ziyun Wang, Yuan Yao, Zhiyuan Liu, and Maosong
  Sun. 2018.
\newblock Fewrel: A large-scale supervised few-shot relation classification
  dataset with state-of-the-art evaluation.
\newblock In \emph{EMNLP}, pages 4803--4809.

\bibitem[{He et~al.(2015)He, Lewis, and Zettlemoyer}]{he2015question}
Luheng He, Mike Lewis, and Luke Zettlemoyer. 2015.
\newblock Question-answer driven semantic role labeling: Using natural language
  to annotate natural language.
\newblock In \emph{EMNLP}, pages 643--653.

\bibitem[{Joshi et~al.(2020)Joshi, Chen, Liu, Weld, Zettlemoyer, and
  Levy}]{joshi2020spanbert}
Mandar Joshi, Danqi Chen, Yinhan Liu, Daniel~S Weld, Luke Zettlemoyer, and Omer
  Levy. 2020.
\newblock Spanbert: Improving pre-training by representing and predicting
  spans.
\newblock \emph{TACL}, 8:64--77.

\bibitem[{Kingma and Ba(2015)}]{kingma2014adam}
Diederik~P Kingma and Jimmy Ba. 2015.
\newblock Adam: A method for stochastic optimization.
\newblock In \emph{ICLR}.

\bibitem[{Kolluru et~al.(2020)Kolluru, Adlakha, Aggarwal, Chakrabarti
  et~al.}]{kolluru2020oie6}
Keshav Kolluru, Vaibhav Adlakha, Samarth Aggarwal, Soumen Chakrabarti, et~al.
  2020.
\newblock Openie6: Iterative grid labeling and coordination analysis for open
  information extraction.
\newblock \emph{arXiv preprint arXiv:2010.03147}.

\bibitem[{Lewis et~al.(2020)Lewis, Liu, Goyal, Ghazvininejad, Mohamed, Levy,
  Stoyanov, and Zettlemoyer}]{lewis2019bart}
Mike Lewis, Yinhan Liu, Naman Goyal, Marjan Ghazvininejad, Abdelrahman Mohamed,
  Omer Levy, Veselin Stoyanov, and Luke Zettlemoyer. 2020.
\newblock Bart: Denoising sequence-to-sequence pre-training for natural
  language generation, translation, and comprehension.
\newblock In \emph{ACL}, pages 7871--7880.

\bibitem[{Li and Liang(2021)}]{li2021prefix}
Xiang~Lisa Li and Percy Liang. 2021.
\newblock Prefix-tuning: Optimizing continuous prompts for generation.
\newblock \emph{arXiv preprint arXiv:2101.00190}.

\bibitem[{Liu et~al.(2021{\natexlab{a}})Liu, Zhang, Hou, Mian, Wang, Zhang, and
  Tang}]{liu2021self}
Xiao Liu, Fanjin Zhang, Zhenyu Hou, Li~Mian, Zhaoyu Wang, Jing Zhang, and Jie
  Tang. 2021{\natexlab{a}}.
\newblock Self-supervised learning: Generative or contrastive.
\newblock \emph{TKDE}.

\bibitem[{Liu et~al.(2021{\natexlab{b}})Liu, Zheng, Du, Ding, Qian, Yang, and
  Tang}]{liu2021gpt}
Xiao Liu, Yanan Zheng, Zhengxiao Du, Ming Ding, Yujie Qian, Zhilin Yang, and
  Jie Tang. 2021{\natexlab{b}}.
\newblock Gpt understands, too.
\newblock \emph{arXiv preprint arXiv:2103.10385}.

\bibitem[{Liu et~al.(2019)Liu, Ott, Goyal, Du, Joshi, Chen, Levy, Lewis,
  Zettlemoyer, and Stoyanov}]{abs-1907-11692}
Yinhan Liu, Myle Ott, Naman Goyal, Jingfei Du, Mandar Joshi, Danqi Chen, Omer
  Levy, Mike Lewis, Luke Zettlemoyer, and Veselin Stoyanov. 2019.
\newblock Roberta: A robustly optimized bert pretraining approach.
\newblock \emph{arXiv preprint arXiv:1907.11692}.

\bibitem[{Mesquita et~al.(2013)Mesquita, Schmidek, and
  Barbosa}]{mesquita2013effectiveness}
Filipe Mesquita, Jordan Schmidek, and Denilson Barbosa. 2013.
\newblock Effectiveness and efficiency of open relation extraction.
\newblock In \emph{EMNLP}, pages 447--457.

\bibitem[{Paolini et~al.(2021)Paolini, Athiwaratkun, Krone, Ma, Achille,
  ANUBHAI, dos Santos, Xiang, and Soatto}]{paolini2021structured}
Giovanni Paolini, Ben Athiwaratkun, Jason Krone, Jie Ma, Alessandro Achille,
  RISHITA ANUBHAI, Cicero~Nogueira dos Santos, Bing Xiang, and Stefano Soatto.
  2021.
\newblock Structured prediction as translation between augmented natural
  languages.
\newblock In \emph{ICLR}.

\bibitem[{Petroni et~al.(2020)Petroni, Lewis, Piktus, Rockt{\"a}schel, Wu,
  Miller, and Riedel}]{petroni2020context}
Fabio Petroni, Patrick Lewis, Aleksandra Piktus, Tim Rockt{\"a}schel, Yuxiang
  Wu, Alexander~H Miller, and Sebastian Riedel. 2020.
\newblock How context affects language models' factual predictions.
\newblock In \emph{AKBC}.

\bibitem[{Petroni et~al.(2019)Petroni, Rockt{\"a}schel, Riedel, Lewis, Bakhtin,
  Wu, and Miller}]{petroni2019language}
Fabio Petroni, Tim Rockt{\"a}schel, Sebastian Riedel, Patrick Lewis, Anton
  Bakhtin, Yuxiang Wu, and Alexander Miller. 2019.
\newblock Language models as knowledge bases?
\newblock In \emph{EMNLP}, pages 2463--2473.

\bibitem[{Radford et~al.(2021)Radford, Kim, Hallacy, Ramesh, Goh, Agarwal,
  Sastry, Askell, Mishkin, Clark et~al.}]{radford2021learning}
Alec Radford, Jong~Wook Kim, Chris Hallacy, Aditya Ramesh, Gabriel Goh,
  Sandhini Agarwal, Girish Sastry, Amanda Askell, Pamela Mishkin, Jack Clark,
  et~al. 2021.
\newblock Learning transferable visual models from natural language
  supervision.
\newblock \emph{arXiv preprint arXiv:2103.00020}.

\bibitem[{Radford et~al.(2018)Radford, Narasimhan, Salimans, and
  Sutskever}]{radford2018improving}
Alec Radford, Karthik Narasimhan, Tim Salimans, and Ilya Sutskever. 2018.
\newblock Improving language understanding by generative pre-training.

\bibitem[{Radford et~al.(2019)Radford, Wu, Child, Luan, Amodei, and
  Sutskever}]{radford2019language}
Alec Radford, Jeffrey Wu, Rewon Child, David Luan, Dario Amodei, and Ilya
  Sutskever. 2019.
\newblock Language models are unsupervised multitask learners.
\newblock \emph{OpenAI Blog}, (8):9.

\bibitem[{Raffel et~al.(2020)Raffel, Shazeer, Roberts, Lee, Narang, Matena,
  Zhou, Li, and Liu}]{raffel2019exploring}
Colin Raffel, Noam Shazeer, Adam Roberts, Katherine Lee, Sharan Narang, Michael
  Matena, Yanqi Zhou, Wei Li, and Peter~J Liu. 2020.
\newblock Exploring the limits of transfer learning with a unified text-to-text
  transformer.
\newblock \emph{JMLR}, 21:1--67.

\bibitem[{Ro et~al.(2020)Ro, Lee, and Kang}]{ro2020multi}
Youngbin Ro, Yukyung Lee, and Pilsung Kang. 2020.
\newblock {M}ulti{\^{}}2{OIE}: Multilingual open information extraction based
  on multi-head attention with {BERT}.
\newblock In \emph{Findings of EMNLP}, pages 1107--1117.

\bibitem[{Schmitz et~al.(2012)Schmitz, Soderland, Bart, Etzioni
  et~al.}]{schmitz2012open}
Michael Schmitz, Stephen Soderland, Robert Bart, Oren Etzioni, et~al. 2012.
\newblock Open language learning for information extraction.
\newblock In \emph{EMNLP}, pages 523--534.

\bibitem[{Shin et~al.(2020)Shin, Razeghi, Logan~IV, Wallace, and
  Singh}]{shin2020autoprompt}
Taylor Shin, Yasaman Razeghi, Robert~L Logan~IV, Eric Wallace, and Sameer
  Singh. 2020.
\newblock Eliciting knowledge from language models using automatically
  generated prompts.
\newblock In \emph{EMNLP}, pages 4222--4235.

\bibitem[{Soares et~al.(2019)Soares, FitzGerald, Ling, and
  Kwiatkowski}]{soares2019matching}
Livio~Baldini Soares, Nicholas FitzGerald, Jeffrey Ling, and Tom Kwiatkowski.
  2019.
\newblock Matching the blanks: Distributional similarity for relation learning.
\newblock In \emph{ACL}, pages 2895--2905.

\bibitem[{Stanovsky and Dagan(2016)}]{stanovsky2016creating}
Gabriel Stanovsky and Ido Dagan. 2016.
\newblock Creating a large benchmark for open information extraction.
\newblock In \emph{EMNLP}, pages 2300--2305.

\bibitem[{Stanovsky et~al.(2016)Stanovsky, Ficler, Dagan, and
  Goldberg}]{stanovsky2016getting}
Gabriel Stanovsky, Jessica Ficler, Ido Dagan, and Yoav Goldberg. 2016.
\newblock Getting more out of syntax with props.
\newblock \emph{arXiv preprint arXiv:1603.01648}.

\bibitem[{Stanovsky et~al.(2018)Stanovsky, Michael, Zettlemoyer, and
  Dagan}]{stanovsky2018supervised}
Gabriel Stanovsky, Julian Michael, Luke Zettlemoyer, and Ido Dagan. 2018.
\newblock Supervised open information extraction.
\newblock In \emph{NAACL}, pages 885--895.

\bibitem[{Wang et~al.(2020)Wang, Liu, and Song}]{wang2020language}
Chenguang Wang, Xiao Liu, and Dawn Song. 2020.
\newblock Language models are open knowledge graphs.
\newblock \emph{arXiv preprint arXiv:2010.11967}.

\bibitem[{Wolf et~al.(2020)Wolf, Chaumond, Debut, Sanh, Delangue, Moi, Cistac,
  Funtowicz, Davison, Shleifer et~al.}]{wolf-etal-2020-transformers}
Thomas Wolf, Julien Chaumond, Lysandre Debut, Victor Sanh, Clement Delangue,
  Anthony Moi, Pierric Cistac, Morgan Funtowicz, Joe Davison, Sam Shleifer,
  et~al. 2020.
\newblock Transformers: State-of-the-art natural language processing.
\newblock In \emph{EMNLP}, pages 38--45.

\bibitem[{Xu et~al.(2013)Xu, Kim, Quinn, Goebel, and Barbosa}]{xu2013open}
Ying Xu, Mi-Young Kim, Kevin~M Quinn, Randy Goebel, and Denilson Barbosa. 2013.
\newblock Open information extraction with tree kernels.
\newblock In \emph{NAACL}, pages 868--877.

\bibitem[{Zhang et~al.(2017)Zhang, Zhong, Chen, Angeli, and
  Manning}]{zhang2017tacred}
Yuhao Zhang, Victor Zhong, Danqi Chen, Gabor Angeli, and Christopher~D.
  Manning. 2017.
\newblock Position-aware attention and supervised data improve slot filling.
\newblock In \emph{EMNLP}, pages 35--45.

\end{thebibliography}
\bibliographystyle{acl_natbib}

\appendix
\newpage
\section{Experimental Setup}
\label{sec:expsetup}

In all experiments, for the generating stage in Sec.~\ref{sec:gen}, we use a pre-trained BERT$_{\rm{LARGE}}$ model~\cite{devlin2018bert} for the beam search. In particular, we use the mean operation over the multi-head attention weights from the last layer of BERT$_{\rm{LARGE}}$ according to the parameter study in \cite{wang2020language}. For the ranking model in Sec.~\ref{sec:gen}, we use the pre-trained BERT$_{\rm{BASE}}$ model. We use the implementations of the pre-trained language models (LM) in the Transformers package~\cite{wolf-etal-2020-transformers}.

To keep our framework simple, we use the same hyperparameters across the majority of our experiments. For generating, we use: \zui{check}8 GeForce RTX 2080 Ti GPUs with a batch size of 16 per GPU; maximum sequence length as 256 tokens. For datasets that provide examples beyond sentence level, we adopt spaCy sentencizer~\footnote{\label{ft:spacysent}{\tiny\url{https://spacy.io/api/sentencizer}}} to segment the texts into sentences, and each sample in the batch is a sentence. For ranking, we experiment with\haoyun{check}: 1 NVIDIA Tesla V100 GPU with a batch size of 8 per GPU; the AdamW optimizer~\cite{kingma2014adam}; linear learning rate decay starting from $10^{-6}$; maximum sequence length as 512 tokens. The top-1 triple from ranking model is returned except for OIE2016 (open information extraction) since it provides a dev set. We additionally report results of top-10 triples for relation classification (FewRel and TACRED).

In the rest of this section, we describe datasets, comparison methods, additional implementation details for each information extraction task, as well as more experimental insights. Results of all experiments are in Table~\ref{tab:allres}. We use the default evaluation metrics for each task as below. We show more input and output formats on all datasets in Table~\ref{tab:inoutexp}.

\subsection{Open Information Extraction}
\paragraph{Datasets} We evaluate the performance of the open information extraction (OIE) systems on OIE benchmark datasets consisting of {\bf OIE2016}~\cite{stanovsky2016creating}, a dataset from Newswire and Wikipedia automatically converted from QA-SRL~\cite{he2015question}; three news datasets {\bf NYT}, {\bf WEB}~\cite{mesquita2013effectiveness}, {\bf PENN}~\cite{xu2013open}. The statistics of the benchmark is shown in Table~\ref{tab:oiedata}. 
\begin{table}[t]
\centering
\resizebox{\linewidth}{!}
  {
  \begin{tabular} {l | l | l | l l l}
        \toprule
    \multirow{2}{*}{{\bf Dataset}} & \multirow{2}{*}{{\bf Domain}} & \multirow{2}{*}{{\bf \#Sents}} & \multicolumn{3}{l}{{\bf \#Triples}} \\
    & & & Train & Dev & Test \\
    \hline
    OIE2016                              & News,Wiki & 3,200 & 5,078 & 1,673 & 1,730 \\   
    WEB                              & News,Web & 500 & - & - & 461 \\   
    NYT                              & News,Wiki & 222 & - & - & 150 \\   
    PENN                              & Mixed & 100 & - & - & 52\\     
    \bottomrule
  \end{tabular}
  }
  \vspace{-0.1in}
\caption{{\small Statistics of OIE benchmark datasets.\zui{add}\xiao{add}}}
\label{tab:oiedata}
\end{table}

\paragraph{Evaluation Methodology} We follow typical OIE metrics to evaluate the systems. First, we report precision, recall, and F1 score using a confidence threshold optimized on the development set. Second, we compute a precision-recall curve by evaluating the performance of the systems at different confidence thresholds. Third, we also measure the area under the PR curve (AUC) to evaluate the overall performance of a system. To compute the above metrics, we need to match the system extractions with the gold extractions. Regarding the matching functions, we adopt the function of OIE2016 to evaluate OIE2016, NYT, WEB, and PENN.

\paragraph{Comparison Methods} We compare our method \method\ to the following prominent OIE systems recently evaluated in ~\cite{stanovsky2018supervised}: ClausIE~\cite{del2013clausie}, Open IE4~\footnote{\label{ft:openie51}\tiny\url{https://github.com/dair-iitd/OpenIE-standalone}}, PropS~\cite{stanovsky2016getting}, RnnOIE~\cite{stanovsky2018supervised}. We also compare to MAMA with BERT$_{\rm{LARGE}}$ recently introduced in \cite{wang2020language} that also leverages pre-trained LMs to extract open triples.

\begin{figure}
\centering
\includegraphics[width=\linewidth]{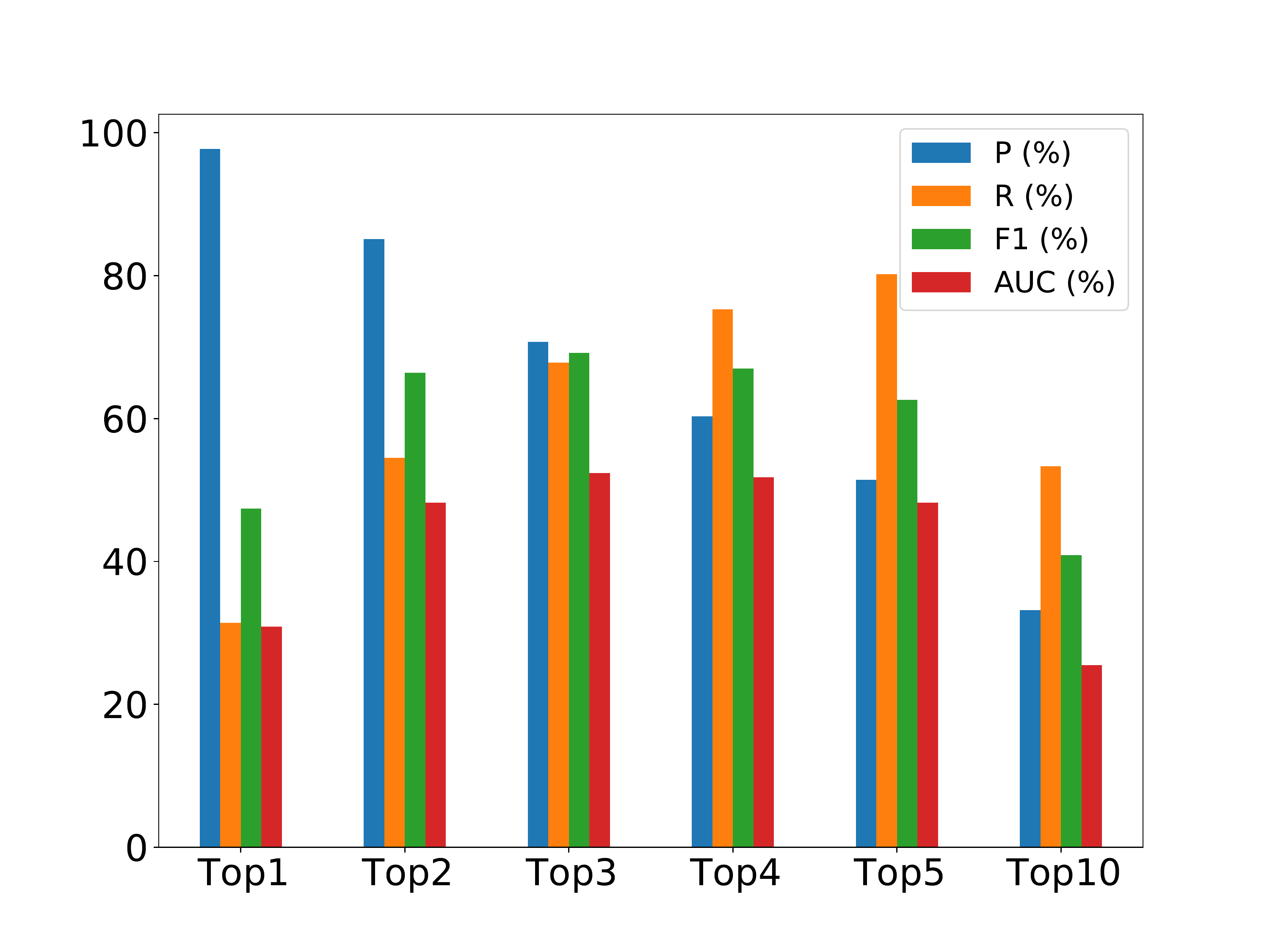}
\caption{{\small Effect of the ranking model on the OIE2016 dev set.}}
\label{fig:cftop}
\end{figure}

\paragraph{Implementation Details} For input text encoding, we use spaCy noun chunks~\footnote{\label{ft:spacynp}\tiny{\url{https://spacy.io/usage/linguistic-features/\#noun-chunks}}} to identify the NPs. We use: beam size equals to 6; the top-1 triple from the ranking model for evaluation except on OIE2016. Instead, we use top-3 triples on OIE2016 based on the parameter study on its dev set in Figure~\ref{fig:cftop}. Experimenting on more OIE benchmark datasets such as CaRB~\cite{bhardwaj2019carb} is an interesting future direction to explore.

\subsection{Relation Classification}
\paragraph{Datasets and Metrics} We evaluate on FewRel~\cite{han2018fewrel} and TACRED~\cite{zhang2017tacred}. 
\begin{itemize}
    \item {\bf FewRel} contains 100 relations with 7 instances for each relation. The standard evaluation for this benchmark uses few-shot N-way K-shot settings. The entire dataset is split into train (64 relations), validation (16 relations) and test set (20 relations). We report the same results on the dev set for all the settings because of our zero-shot setting.
    \item {\bf TACRED} is a large-scale relation classification benchmark that consists of 106,344 examples and 41 relation types including 68,164 for training, 22,671 for validation, and 15,509 for testing. We do not use train and validation sets, and report the result on the test set.
\end{itemize}
We use F1 to evaluate the results.

\paragraph{Comparison Methods} We compare our method with the following supervised methods. (\expandafter{\romannumeral1}) BERT-PAIR~\cite{gao2019fewrel} is a sequence classification model based on BERT, optimizing the score of two instances expressing the same relation. (\expandafter{\romannumeral2}) BERT$_{\rm EM}$ + Matching the Blanks (MTB)~\cite{soares2019matching}, which uses entity markers (BERT$_{\rm EM}$) and additional pre-training of relations on a large-scale corpus (i.e., MTB). (\expandafter{\romannumeral3}) TANL~\cite{paolini2021structured} is a sequence to sequence model based on T5~\cite{raffel2019exploring} aiming to generate structured objects from an encoded natural language format. 

\paragraph{Implementation Details} For input text encoding, we attach the given gold head and tail entities to the input. As described in Sec.~\ref{sec:rc}, linked relation phrases are also attached to the input. For TACRED, we use the relation map provided in \cite{angeli2015leveraging} to link relation phrases to TACRED relations, and manually build a map between TACRED relations and Wikidata predicates. For FewRel, we directly link relation phrases to gold Wikidata predicates as FewRel uses Wikidata predicates as the target category. At test time, \method\ makes prediction as below: given a returned triple, we first map the relation phrase of the triple back to a Wikidata predicate, then use the Wikidata predicate to find a relation in the pre-defined category. We report the results using two setups: top-1 and top-10 triples from the ranking model. In practice, some of the Wikidata predicates do not have aliases. We therefore follow \cite{angeli2015leveraging,wang2020language} to manually add aliases for such predicates based on the alignment between Wikipedia and Wikidata. We set the beam size as 6\haoyun{check}.

\begin{table*}[]
\resizebox{\linewidth}{!}{
\begin{tabular}{@{}llp{16cm}@{}}
\toprule
{\bf Error type}                                & {\bf Percentage}          & {\bf Example (sentence and gold triple)}                                                \\ \midrule
\multirow{2}{*}{Wrong NP} & \multirow{2}{*}{46\%} & {\sl San Jose , Calif. - based Adobe announced in April the plans to {\bf acquire} Macromedia , which makes the Flash animation software used to display graphics 
on Web sites .} \\
                                          &                     & (Adobe; {\bf acquire}; Macromedia)                                          \\ \midrule
\multirow{2}{*}{Unformatted Sentence} & \multirow{2}{*}{14\%} & {\sl Adobe {\bf acquired} Macromedia! - sephiroth.it - flash \&amp;amp; php} \\
                                          &                     & (Adobe; {\bf acquired}; Macromedia!)                                          \\ \midrule
\multirow{2}{*}{Nominalization} & \multirow{2}{*}{12\%} & {\sl Google confirms YouTube {\bf acquisition} - BBC News} \\
                                          &                     & (Google; {\bf acquisition}; YouTube)                                          \\ \midrule
\multirow{2}{*}{Long Sentence} & \multirow{2}{*}{10\%} & {\sl Amid all the hubbub over Google 's {\bf swallowing} of YouTube , we ve heard both considered commentary and over - the - top pontification about whether it's a good deal or a bad deal , for the newly - engorged company and for all those users in TV land .} \\
                                          &                     & (Google; {\bf swallowing}; YouTube)                                          \\ \midrule
\multirow{2}{*}{Noun}                     & \multirow{2}{*}{4\%} & {\sl Dean Kamen ( left ) , {\bf inventor} of the Segway Human Transporter Human Transporter wearing the Plantronics Voyager 510 Bluetooth Headset}    \\
                                          &                     & (Dean Kamen; {\bf inventor}; the Segway Human Transporter)                                               \\ \bottomrule
\end{tabular}}
\caption{{\small Analysis of frequently-occurring recall errors of \method\ on a random sample of 50 sentences on the OIE task. For each type we list the percentage of sentences in which it occurs, and an example taken from the WEB corpus.}\zui{add}\xiao{add}}
\label{tab:errmain}
\end{table*}
\begin{table*}[]
\resizebox{\linewidth}{!}{
\begin{tabular}{@{}lp{16cm}lll@{}}
\toprule
{\bf Dataset}                    & {\bf Example (sentence and gold triple)}                                                                                                                           & {\bf LAMA-Oracle}         & {\bf \method} (\electricblue{\small ours}) & {\bf Error Type}            \\ \midrule
\multirow{4}{*}{T-REx}     & {\sl Naomi Shihab Nye (born March 12, 1952) is a {\bf poet}, songwriter, and novelist.} & \multirow{2}{*}{teacher \xmark}     & \multirow{2}{*}{poet \cmark} & \multirow{2}{*}{Wrong memory} \\
                           & (Naomi Shihab Nye; occupation; {\bf poet})                                                                                                             &                       &                    \\ \cmidrule(l){2-5}
                           & {\sl Nicholas Liverpool died on 1 June 2015 in {\bf Miami}, where he was receiving medical treatment.} & \multirow{2}{*}{London \xmark}     & \multirow{2}{*}{Miami \cmark} & \multirow{2}{*}{Wrong memory} \\
                           & (Nicholas Liverpool; place\_of\_death; {\bf Miami})                                                                                                             &                       &                    \\\cmidrule(l){2-5}
                           & {\sl Jean-Michel Pilc (born 1960 in Paris, France) is a self-taught {\bf jazz} pianist currently residing in New York.} & \multirow{2}{*}{classical \xmark}     & \multirow{2}{*}{jazz \cmark} & \multirow{2}{*}{Wrong memory}  \\
                           & (Jean-Michel Pilc; genre; {\bf jazz})                                                                                                             &                       &                    \\\cmidrule(l){2-5}
                           & {\sl Nick Lucas's version, released on {\bf Brunswick}, was a No.} & \multirow{2}{*}{EMI \xmark}     & \multirow{2}{*}{Brunswick \cmark} & \multirow{2}{*}{Wrong memory}  \\
                           & (Nick Lucas; record\_label; {\bf Brunswick})                                                                                                             &                       &                    \\
                           \bottomrule
\end{tabular}}
\caption{{\small Out-of-context predictions of LAMA-Oracle~\cite{petroni2020context} for the factual probing task. \xiao{add}}}
\label{tab:lamamistake}
\end{table*}
\begin{table*}[]
\resizebox{\linewidth}{!}{
\begin{tabular}{@{}lp{8cm}@{\hskip 0.4in}p{8cm}@{}}
\toprule
{\bf Dataset}          & {\bf Input}                                                                                                                          & {\bf Output}                                                                                                                         \\ \midrule
OIE2016          & \underline{He}$_{\rm NP}$ hasn't been able to replace \underline{the M'Bow cabal}$_{\rm NP}$ .                                                     & (He; hasn't been able to replace; the M'Bow cabal)                                                                                                                               \\ \hline
WEB              &   \underline{Crick}$_{\rm NP}$ received \underline{a Nobel Prize}$_{\rm NP}$ for discovering \underline{the structure}$_{\rm NP}$ of \underline{DNA}$_{\rm NP}$ .                                                                                                                            &  (Crick; received a Nobel Prize for; the structure), (DNA; for discovering the structure; a Nobel Prize)                                                                                                                         \\ \hline
NYT              & PRODUCER -- \underline{Squier Knapp Dunn Communications}$_{\rm NP}$ in consultation with \underline{David Garth}$_{\rm NP}$                            & (David Garth; in consultation with; Squier Knapp Dunn Communications)                                                                                                                      \\ \hline
PENN             &  \underline{A spokeswoman}$_{\rm NP}$ said \underline{Sulka}$_{\rm NP}$ operates \underline{a total}$_{\rm NP}$ of seven stores in \underline{the U.S.}$_{\rm NP}$ and overseas .                                                                                                                            &  (Sulka; operates a total of; the U.S), (A spokeswoman; said; the U.S), (a total; of seven stores; the U.S)                                                                                                               \\ \hline
TACRED           &  \underline{Denise Maloney Pictou}$_{\rm GOLD}$, one of \underline{Aquash}$_{\rm GOLD}$'s \underline{daughters}$_{\rm child}$, says she hopes Graham 's trial will help bring justice to her family .                                                                                                                         &       (Denise Maloney Pictou; child; Aquash)                                                                                                                         \\ \hline
FewRel 1.0 (dev) &  \underline{Theodore II Palaiologos}$_{\rm GOLD}$ was a son of the Eastern Roman Emperor Manuel II Palaiologos and his \underline{wife}$_{\rm spouse}$ \underline{Helena Draga}$_{\rm GOLD}$ .                                                                                                                              &         (Theodore II Palaiologos; spouse; Helena Draga)                                                                                                                       \\\hline
Google-RE        &  \underline{Peter F Martin}$_{\rm GOLD/NP}$ (\underline{born}$_{\rm birth\_date}$ \underline{1941}$_{\rm NP}$) is an \underline{American politician}$_{\rm NP}$ who is a \underline{Democratic member}$_{\rm NP}$ of the \underline{Rhode Island House of Representatives}$_{\rm GOLD/NP}$ . & (Peter F Martin; ${\rm date\_of\_birth}$; 1941) \\ \hline
T-REx            & \underline{Antonio Agliardi}$_{\rm GOLD/NP}$ in 4 September 1832 – 19 March 1915 \underline{was an Italian}$_{\rm position\_held}$ \underline{Roman Catholic Cardinal}$_{\rm NP}$ , \underline{archbishop}$_{\rm NP}$ , and \underline{papal diplomat}$_{\rm NP}$ .                                                                                                                               & (Antonio Agliardi; ${\rm position\_held}$; diplomat)                                                                                                                              \\ 
\bottomrule
\end{tabular}}
\caption{{\small Input/output examples for all datasets.}\zui{add}\haoyun{add}\xiao{add}}
\label{tab:inoutexp}
\end{table*}

\subsection{Factual Probe}
\paragraph{Datasets and Metrics} We consider the Google-RE consisting of 3 relations and 5,527 facts, and T-REx with 41 relations and 34,039 facts of the LAMA benchmark~\cite{petroni2019language}. We evaluate the results using mean precision at one (P@1), where higher values are better. 

\paragraph{Comparison Methods} We compare to pre-trained LM based methods that leverage the output probabilities of the LM to make predictions given the sentence known to express the fact. Two methods are considered: (\expandafter{\romannumeral1}) LAMA~\cite{petroni2019language} leverages the input sentence without the tail entity to query the LMs, and (\expandafter{\romannumeral2}) LAMA-Oracle~\cite{petroni2020context} enriches the query with (at most) five gold sentences as additional context. 

\paragraph{Implementation Details} For input encoding, we use spaCy noun chunks~\footref{ft:spacynp} to identify the NPs. We additionally use Stanford NER~\footnote{{\tiny\url{https://stanfordnlp.github.io/CoreNLP/ner.html}}} to label the missing entities from noun chunks such as dates. A NP is considered as a gold head entity if it overlaps with the mention of the gold head entity. Therefore there might be multiple NPs representing the gold head entity in the input. Similar to relation classification, linked relation phrases are attached to the input. We use the gold mapping to convert Freebase predicates to Wikidata ones for Google-RE, and use the gold Wikidata predicates for T-REx as T-REx builds based on Wikidata. At test time, a prediction is made: given a returned triple, we conduct an exact match between the tail entity of the triple and the gold tail entity. We use the top-1 returned triple for P@1 evaluation. The beam size equals 20\xiao{check}.

\section{Additional Analysis of Results}
\label{sec:aar}

\paragraph{Error Analysis of OIE} We show detailed error analysis of \method\ on the OIE task in Table~\ref{tab:errmain}.

\paragraph{Relation Position Distribution} We show the statistics of relation positions in the gold triples in all OIE datasets. Overall, there is a considerable amount of triples containing relation not between the entity pair across all OIE datasets as shown in Table~\ref{tab:statrel}, demonstrating the importance of triple-oriented beam search. In particular, 14.8\% of the triples contains relations outside the entity pairs on OIE datasets (OIE2016, WEB, NYT, PENN). 
\begin{table}[]
\resizebox{\linewidth}{!}{
\begin{tabular}{@{}lllll@{}}
\toprule
\multirow{2}{*}{{\bf Dataset}} & \multicolumn{4}{l}{{\bf \#Triples}}     \\
                         & {\bf Left Relation} & {\bf Right Relation} & {\bf Middle Relation}     & {\bf Total} \\ \midrule
OIE2016                  & 128  & 165   & 1,437 & 1,730  \\
WEB                      & 17   & 29    & 415        & 461   \\
NYT                      & 2    & 29    & 119        & 150   \\
PENN                     & 2    & 4     & 46         & 52    \\ \bottomrule
\end{tabular}}
\caption{{\small Statistics of relation positions of the gold triples in all OIE datasets.\haoyun{add}\xiao{add}\todo{make relation in the left, right in the title: clearer?}}}
\label{tab:statrel}
\end{table}

\paragraph{Output Statistics} Table~\ref{tab:outstats} compares the statistics of the outputs of different OIE systems and gold data on OIE2016. We find that \method\ produces 25 more triples than the gold data, and the arguments tend to be shorter.
\begin{table}[]
\centering
\resizebox{0.6\linewidth}{!}{%
\begin{tabular}{@{}lcc@{}}
\toprule
{\bf System}                 & {\bf \#Triples} & {\bf Words/Arg} \\ \midrule
Gold                   & 1,730      & 5.38     \\
ClausIE                & 2,768      & 5.78     \\
Open IE4               & 1,793      & 4.55     \\
PropS                  & 1,551      & 5.80     \\
RnnOIE                 & 1,993      & 4.68      \\
MAMA                   & 1751 & 2.56 \\
\method\ (\electricblue{\small ours}) & 1755 & 2.30\\ \bottomrule
\end{tabular}%
}
\caption{{\small Output statistics of the different OIE systems on OIE2016.\zui{add}}}
\label{tab:outstats}
\end{table}

\paragraph{Out-of-Context Cases for LAMA-Oracle} Table~\ref{tab:lamamistake} shows major error cases in LAMA-Oracle when compared with \method, where LAMA-Oracle often generates out-of-context answers due to the wrong memory of LMs. This shows that explainable extraction like \method\ is necessary for information extraction using LMs.

\section{Analysis of Ranking Model}
\paragraph{Implementation Details} \haoyun{check} We remove triples with empty predicates from the original T-REx~\cite{elsahar2019t}, and further remove sentence-triple pairs if the triples are in the LAMA version of T-REx~\cite{petroni2019language}. This results in approximately 4 million positive sentence-triple pairs. We set the number of fine-tuning epochs to 1. Each batch takes approximately 7 minutes, resulting in a total 12 hours for fine-tuning the ranking model.

\begin{figure}
\centering
\includegraphics[width=\linewidth]{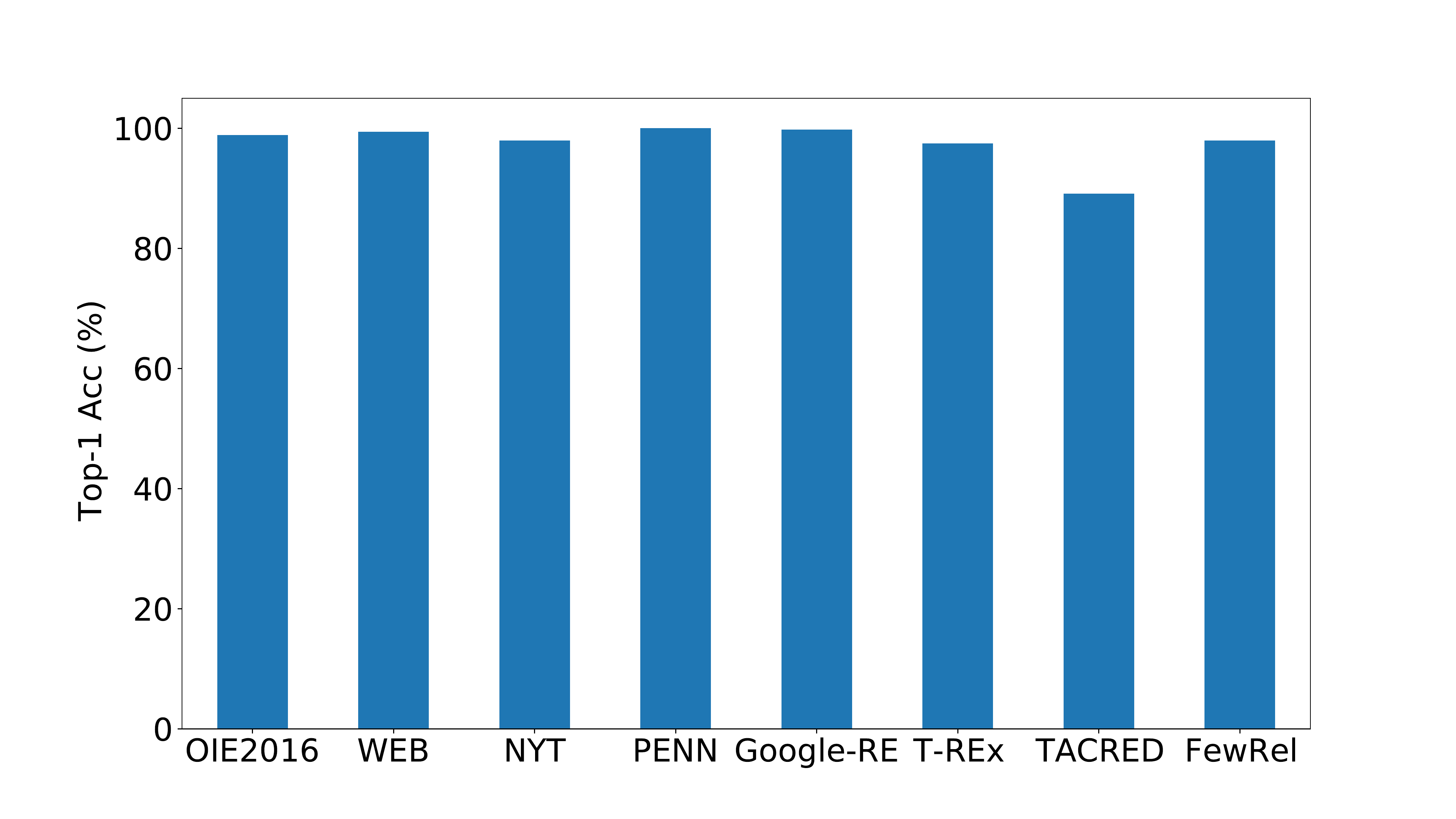}
\caption{{\small Top-1 accuracy of the ranking on all datasets.\haoyun{update}}}
\label{fig:cfres}
\end{figure}

\paragraph{Ranking Results} We evaluate the performance of the ranking model. To do so, we construct negative sentence-triple samples for all the tested datasets as follows: for each sentence in a dataset, we randomly sample a triple from other sentences to construct a negative sample. We then directly evaluate the ranking performance of the model on all the datasets, and report the top-1 accuracy in Figure~\ref{fig:cfres}. We find that the ranking model works extremely well, obtaining nearly perfect top-1 accuracy on all the datasets.

\end{document}